\documentclass[times,10pt,twocolumn]{article}
\usepackage{wacv}
\usepackage{times}
\usepackage{epsfig}
\usepackage{graphicx}
\usepackage{amsmath}
\usepackage{amssymb}
\usepackage{booktabs}

\usepackage{cuted}
\usepackage[ruled,vlined]{algorithm2e}
\usepackage{algpseudocode}
\usepackage{tabularx}
\usepackage{subcaption}
\usepackage{lipsum}
\usepackage{multirow}
\usepackage{multicol}
\usepackage{pifont}
\usepackage{xcolor}
\usepackage[numbers,sort]{natbib}
\usepackage[inline]{enumitem}
\usepackage{pgfkeys}
\usepackage[normalem]{ulem}
\usepackage{authblk}

\usepackage{tikz}
\usetikzlibrary{shapes.geometric,arrows,positioning}
\tikzstyle{startstop} = [rectangle, rounded corners, minimum width=2cm, minimum
height=1cm,text centered, draw=black, fill=red!30]
\tikzstyle{process} = [rectangle, minimum width=2cm, minimum height=1cm, text
centered, draw=black, fill=orange!30]
\tikzstyle{decision} = [diamond, minimum width=2cm, minimum height=1cm, text
centered, draw=black, fill=green!30]
\tikzstyle{io} = [rectangle, minimum width=2cm, minimum height=1cm, text
centered, draw=black, fill=blue!30]
\tikzstyle{arrow} = [thick,->,>=stealth]

\usepackage{pgfplots}
\pgfplotsset{compat=1.12}

\def\wacvPaperID{1397} 

\wacvalgorithmstrack   

\wacvfinalcopy 

\usepackage[pagebackref=true,breaklinks=true,colorlinks,bookmarks=false]{hyperref}

\usepackage{orcidlink}

\begin{document}

\newcommand{\budget}{B}
\newcommand{\pamount}{M}

\newcommand{\fulldata}{\mathcal{D}}
\newcommand{\labeled}{\mathcal{L}}
\newcommand{\unlabeled}{\mathcal{U}}
\newcommand{\pseudo}{\mathcal{P}}

\newcommand{\frames}{F}
\newcommand{\view}{V}
\newcommand{\numviews}{N}
\newcommand{\kp}{k}
\newcommand{\totalkp}{K}

\newcommand{\heatmap}{H}
\newcommand{\localpeaks}{L}
\newcommand{\peak}{l}
\newcommand{\xyp}{\peak_1}
\newcommand{\xyzp}{P}
\newcommand{\axyzp}{\hat{P}}
\newcommand{\xyzgt}{P^*}
\newcommand{\numin}{c}
\newcommand{\terror}{\varepsilon}

\newcommand{\metric}{\mathcal{M}}
\newcommand{\bsb}{\metric_\text{BSB}}
\newcommand{\mpe}{\metric_\text{MPE}}
\newcommand{\coreset}{\metric_\text{CS}}
\newcommand{\mc}{\metric_\text{MC}}
\newcommand{\random}{\textsc{Rand}}
\newcommand{\BSB}{\textsc{BSB}}
\newcommand{\MPE}{\textsc{MPE}}
\newcommand{\CS}{\textsc{Ours-CS}}
\newcommand{\MC}{\textsc{Ours-MC}}
\newcommand{\sal}{\textsc{ST}}

\newcommand{\panoptic}{CMU Panoptic}
\newcommand{\ih}{InterHand2.6M}
\newcommand{\kun}[1]{{\color{red}(Kun: #1)}}
\newcommand{\fred}[1]{}
\newcommand{\cem}[1]{{\color{green}(Cem: #1)}}
\newcommand{\remove}[1]{}

\definecolor{C1}{RGB}{252,202,108}
\definecolor{C2}{RGB}{200,91,108}
\definecolor{C3}{RGB}{49,54,88}
\definecolor{C4}{RGB}{242,163,94}

\pgfplotscreateplotcyclelist{default}{
  mark=x, color=red, smooth, error bars/.cd, y fixed, y dir=both, y explicit\\
  mark=x, color=blue, smooth, error bars/.cd, y fixed, y dir=both, y explicit\\
  mark=x, color=cyan, smooth, error bars/.cd, y fixed, y dir=both, y explicit\\
  mark=x, color=violet, smooth, error bars/.cd, y fixed, y dir=both, y explicit\\
  mark=x, color=C3, smooth, error bars/.cd, y fixed, y dir=both, y explicit\\
  mark=x, color=C4, smooth, error bars/.cd, y fixed, y dir=both, y explicit\\
  mark=x, color=C1, smooth, error bars/.cd, y fixed, y dir=both, y explicit\\
  mark=x, color=C2, smooth, error bars/.cd, y fixed, y dir=both, y explicit\\
}

\pgfplotscreateplotcyclelist{heatmap}{
  fill=red!90!white\\
  fill=red!70!white\\
  fill=red!50!white\\
  fill=red!30!white\\
  fill=red!10!white\\
  fill=blue!10!white\\
  fill=blue!30!white\\
  fill=blue!50!white\\
  fill=blue!70!white\\
  fill=blue!90!white\\
}

\pgfplotscreateplotcyclelist{cluster}{
  fill=cyan!30!white\\
  fill=cyan!70!white\\
  fill=red!90!white\\
  fill=red!30!white\\
  fill=violet!90!white\\
  fill=violet!40!white\\
  fill=green!50!white\\
  fill=blue!40!white\\
  fill=C2!90!white\\
  fill=C3!30!white\\
}

\usepgfplotslibrary{fillbetween}
\newenvironment{customlegend}[1][]{%
  \begingroup
  \pgfplots@init@cleared@structures
  \pgfplotsset{#1}%
}{%
  \pgfplots@createlegend
  \endgroup
}%

\def\addlegendimage{\pgfplots@addlegendimage}

\author{Qi Feng\orcidlink{0000-0001-6342-3228}}
\author{Kun He\orcidlink{0000-0002-0828-0794}}
\author{He Wen\orcidlink{0000-0002-7788-6896}}
\author{Cem Keskin}
\author{Yuting Ye\orcidlink{0000-0003-2643-7457}}
\affil{
Meta Reality Labs\\
{\tt\small \{fung,kunhe,hewen,cemkeskin,yuting.ye\}@meta.com}
}

\title{Rethinking the Data Annotation Process for Multi-view 3D Pose Estimation\\with Active Learning and Self-Training}
\maketitle

\begin{abstract}
Pose estimation of the human body and hands is a fundamental problem in computer
vision, and learning-based solutions require a large amount of annotated data.
In this work, we improve the efficiency of the data annotation process for 3D
pose estimation problems with Active Learning (AL) in a multi-view setting. AL
selects examples with the highest value to annotate under limited annotation
budgets (time and cost), but choosing the selection strategy is often
nontrivial. We present a framework to efficiently extend existing
single-view AL strategies. We then propose two novel AL strategies that make
full use of multi-view geometry. Moreover, we demonstrate additional performance
gains by incorporating pseudo-labels computed during the AL process, which is a
form of self-training. Our system significantly outperforms simulated annotation
baselines in 3D body and hand pose estimation on two large-scale benchmarks: CMU
Panoptic Studio and InterHand2.6M. Notably, on CMU Panoptic Studio, we are able
to reduce the turn-around time by 60\% and annotation cost by 80\% when compared
to the conventional annotation process.
\end{abstract}

\section{Introduction}
\label{sec:intro}

Pose estimation is a fundamental problem in computer vision. Accurate pose
estimations of the human body/hands allow automated systems to perform markerless
motion capture~\cite{elhayek2015efficient,tome2018rethinking}, recognize
actions~\cite{cheron2015p,yao2011does}, understand social
interactions~\cite{joo2017panoptic} and sign languages~\cite{isaacs2004hand},
and so on.

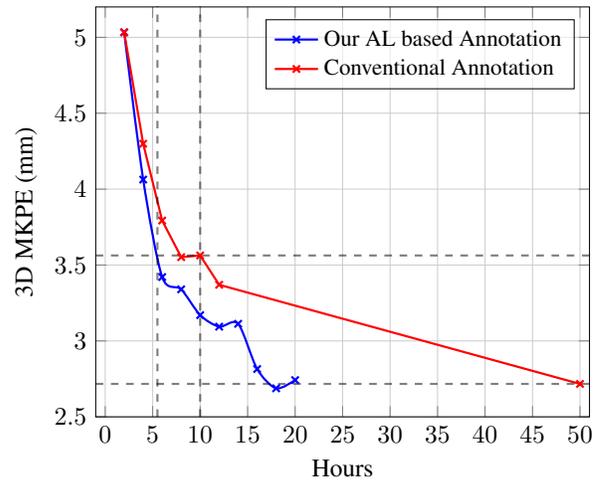
\begin{figure}[!t]
    \centering
    \resizebox{0.95\columnwidth}{!}{
      \begin{tikzpicture}
        \begin{axis}[
            grid=both,
            grid style={line width=.1pt, draw=gray!40},
            ylabel= 3D MKPE (mm),
            xlabel= Hours,
            legend pos = north east,
            legend style={font=\small, fill=white, fill opacity=0.6, draw opacity=1, text opacity=1},
            legend cell align={left},
            cycle list name=default,
            scaled x ticks = false,
            ymin=2.5,ymax=5.2,
            xmin=-1,xmax=51,
            xtick={0,5,...,50},
          ]
          \addplot[mark=x, color=blue, smooth, line width=0.3mm] table[x={HOURS},y={OURS}]{data/perf-time-ours.txt};
          \addlegendentry{Our AL based Annotation};
          \addplot[mark=x, color=red, line width=0.3mm] table[x={HOURS},y={RANDOM}]{data/perf-time-random.txt};
          \addlegendentry{Conventional Annotation};
          \addplot[color=black, dashed, opacity=0.5, line width=0.3mm, domain=-1:51]{2.717};
          \addplot[color=black, dashed, opacity=0.5, line width=0.3mm, domain=-1:51]{3.563};
          \addplot[color=black, dashed, opacity=0.5, line width=0.3mm] coordinates {(5.5, -1) (5.5, 6)};
          \addplot[color=black, dashed, opacity=0.5, line width=0.3mm] coordinates {(10, -1) (10, 6)};
        \end{axis}
      \end{tikzpicture}
    } 
    \caption{ 
    Model test accuracy vs. annotation turn-around time. We use the estimate of 1
    minute per frame for annotation and 1 hour for training the model.
    Conventional Annotation does not require training during annotation. Our
    AL based annotation system saves the overall turn-around time by 45\% for
    PoseResNet with a test performance of 3.5 mm, and more than 60\% with a test
    performance of 2.7mm. 
    \fred{If possible, update the red curve.}
    }
    \vspace{-0.2in}
    \label{fig-perf-vs-time}
  \end{figure}
  

While supervised learning methods using deep neural networks have achieved
considerable success for pose
estimation~\cite{newell2016stacked,wang2020deep,wei2016convolutional,xiao2018simple},
the annotation of pose data is time-consuming and costly.  For example, the
creators of MPII~\cite{andriluka20142d}, a popular body pose estimation
benchmark, reported that on average it takes an annotator \emph{one minute} to
annotate all body keypoints on an image.  Human labels can also have
inconsistent quality, especially for the difficult occluded cases.  On the other
hand, multi-view camera
systems~\cite{joo2017panoptic,yu2020humbi,zimmermann2019freihand} are
increasingly being used to generate pose labels automatically, which is a major
motivation for our work.  However, training the underlying labeling
models still requires significant upfront annotation.

\begin{figure*}
\centering
\resizebox{0.95\textwidth}{!}{
  \begin{tikzpicture}
    \node(gather) at (0,0) {Gathered Raw Data};
    \node[draw, 
      below=of gather,
      rounded corners=0.05 in, 
      minimum height=0.5 in,
    ](unlabeled-pool) at (0,0) {Unlabled Pool};
    \node[draw,
      align=center, 
      right=of unlabeled-pool,
      minimum height=0.5 in, 
      xshift=-0.2in,
    ] (inference) {Inference using \\ Trained Model};
    \node[draw,
      align=center, 
      right=of inference,
      minimum height=0.5 in,
      xshift=0.6in,
    ] (al) {Active Learning (AL) \\ and \\ Self Training (ST)};
    \node[draw,
      align=center, 
      right=of al, 
      xshift=0.6in,
      minimum height=0.5 in,
      minimum width=1 in,
    ] (annotation) {Human \\ Annotation};
    \node[draw,
      align=center, 
      above=of annotation,
      minimum height=0.5 in,
      minimum width=1 in,
    ] (augmentation) {Data\\ Augmentation};
    \node[draw,
      align=center,
      right=of augmentation,
      minimum height=0.5 in,
      xshift=0.8in
    ] (training) {Train Pose\\Estimation Model};
    \node[draw,
      below=of training,
      rounded corners=0.05 in, 
      minimum height=0.5 in,
    ](labeled-pool){Labeled Pool};
    \node[
        right=of labeled-pool,
        xshift=0.3in
    ](labeled){Annotated Data};
    \node[
        right=of training,
    ](model){Pose Estimation Model};

    \draw[arrow] (gather) -- (unlabeled-pool);
    \draw[arrow] (unlabeled-pool) -- (inference);
    \draw[arrow] (inference) -- node[sloped,anchor=center,above] {Pseudo-Labels} (al);
    \draw[arrow] (al) -- node[sloped,anchor=center,above,align=center] {ST: Good \\ Pseudo-Labels} (augmentation);
    \draw[arrow] (al) -- node[sloped,anchor=center,below,align=center] {AL: Bad \\ Pseudo-Labels} (annotation);
    \draw[arrow] (augmentation) -- node[sloped,anchor=center,above,align=center] {Pseudo-Labels} (training);
    \draw[arrow] (annotation) -- node[sloped,anchor=center,below,align=center] {Add to} node[sloped,anchor=center,above,align=center] {Frames removed \\ from unlabeled pool} (labeled-pool);
    \draw[arrow] ([yshift=-0.4 in]labeled-pool) -- node[anchor=center,right,align=center] {Ground Truth\\Labels} (training);    
    \draw[arrow] (training) -- ([yshift=0.1in]training.north) node[sloped,anchor=center,below,align=center,xshift=-4in] {Trained Model}   -| (inference);
    \draw[arrow] (labeled-pool) -- (labeled);
    \draw[arrow] (training) -- (model);
  \end{tikzpicture}
}

\caption{
  An overview of the proposed active learning (AL) system for multi-view 3D
  pose estimation. While prior works have only considered AL for single-view
  pose estimation, our system is the first to work in the multi-view setting
  (Sec.~\ref{sec:AL}), and we propose two effective strategies that make full
  use of multi-view geometry. Additionally, by incorporating pseudo-labels in
  the proposed self-training process (Sec.~\ref{sec:SAL}), we show further
  improvement in annotation efficiency without extra annotation or computational
  cost. 
}
\vspace{-0.2in}
\label{fig-overall}
  \end{figure*}
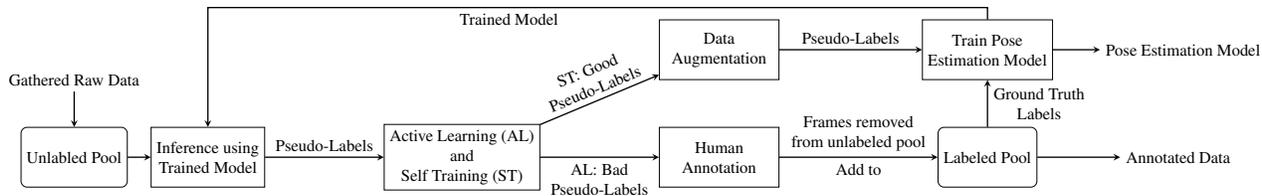

In this paper, we propose an annotation process based on Active Learning
(AL)~\cite{cohn1994improving,settles2009active} to make the data annotation
process for learning deep pose estimation models faster and more
cost-effective.  Our AL based approach focus the annotation budget (time and
cost) on the most valuable samples.

We study AL formulations in the context of pose estimation; in particular, we
consider \emph{3D body and hand pose estimation from multi-view RGB images}.
By exploiting the use of multi-view geometry, we propose two novel AL
strategies that are geometrically inspired, and easy to compute.  To our
knowledge, other existing AL systems for pose
estimation~\cite{liu2017active,caramalau2021active,yoo2019learning} do not
consider multi-view input, and the proposed single-view strategies therein do
not generalize well to the multi-view setup.

In addition to the main AL formulation, we also explore further improvements to
the annotation efficiency with \emph{self-training}, which has been a successful
strategy for image
classification~\cite{xie2020self,zoph2020rethinking,pham2021meta}.  To this
end, during each AL iteration we augment the human-annotated labels with
pseudo-labels computed from the model's prediction, inspired by the multi-view
bootstrapping method~\cite{simon2017hand}.  Our experiments show that, with
careful selection, pseudo-labels can further boost pose estimation performance
without additional annotation or computational cost.

We conduct annotation simulations, experiments, and ablation studies on two
large-scale benchmarks, \panoptic~\cite{joo2017panoptic} for body pose
estimation, and \ih~\cite{moon2020interhand} for hand pose estimation.  Our
proposed multi-view AL strategies, together with the self-training strategies,
consistently outperform baseline strategies by significant margins. Notably, as
shown in Fig.~\ref{fig-perf-vs-time}, on \panoptic{} our complete system reduces the
annotation turn-around time by 60\% and the annotation labor cost by 80\% when
compared to existing data annotation processes.
In summary, our contributions in this paper are threefold:
\begin{itemize}
  \item
    We propose a data annotation process based on Active Learning for 3D pose
    estimation from multi-view RGB images, and propose AL strategies that utilize
    multi-view geometry to reduce the annotation time and cost.
  \item
    We explore self-training for pose estimation in the proposed AL framework,
    and show that further gains can be realized by including pseudo-labels.
  \item
    We show that the proposed AL and self-training strategies significantly
    improve annotation efficiency over baselines, and establish the
    state-of-the-art in AL for multi-view pose estimation.
\end{itemize}


\section{Related Work}\label{sec:related}

\noindent\textbf{3D Pose Estimation}: 
Pose estimation is one of the fundamental tasks in computer vision.  To model
human bodies that can undergo articulation and deformation, early approaches
mostly take inspiration from the classical Pictorial
Structures~\cite{andriluka2009pictorial,felzenszwalb2005pictorial}. Following
the success of deep neural networks, and facilitated by benchmarks such as Human
3.6M~\cite{ionescu2013human3} and MPII~\cite{andriluka20142d}, deep CNNs have
been widely applied to body and hand pose estimation. Representative methods
include Convolutional Pose Machines~\cite{wei2016convolutional}, Stacked
Hourglass Networks~\cite{newell2016stacked}, PoseResNet~\cite{xiao2018simple},
HRNet~\cite{wang2020deep}, \etc. These methods typically work by predicting the
locations of body/hand keypoints, formulated as a heatmap regression problem.
Single-view 3D pose estimation
methods~\cite{martinez2017simple,kanazawa2018end,kolotouros2019learning,zimmermann2019freihand}
on the other hand, directly lift 2D image evidence into 3D keypoints or mesh
representations, but need more high-quality training data in order to resolve
the inherent 2D-3D ambiguity.

With the increasing availability of multi-camera setups, \emph{multi-view} pose
estimation has gathered increased
interest~\cite{iskakov2019learnable,He_2020_CVPR,Remelli_2020_CVPR}. A key
motivation is that these systems can be used to automatically or
semi-automatically generate ``ground truth'' for single-view 3D pose estimation,
and significantly reduce labeling cost. In fact, such a procedure has been
adopted in benchmarks like \panoptic~\cite{joo2017panoptic} and
HUMBI~\cite{yu2020humbi} for body pose estimation, as well as
FreiHand~\cite{zimmermann2019freihand} and \ih~\cite{moon2020interhand} for
hands. However, training multi-view models still requires large amounts of
annotated 3D pose data, which strongly motivates cost-saving strategies such as
active learning.

\noindent\textbf{Active Learning}: 
Active Learning (AL) \cite{cohn1994improving,settles2009active} considers a
dynamic environment where an ML system selects unlabeled examples to acquire
labels for, and iteratively re-trains itself using newly labeled data. This is
critical in many real-world scenarios  with  constrained annotation budgets. A
large AL literature exists for classification, including uncertainty-based
sampling \cite{roth2006margin}, diversity maximization \cite{yang2015multi},
Bayesian methods \cite{sener2018active}, \etc. Despite years of progress, in
practice, the best AL strategies are often problem-dependent, and heuristics
such as random sampling remain strong
baselines~\cite{mittal2019parting,shui2020deep}.

In computer vision, AL has also been widely studied for problems such as
semantic segmentation~\cite{viewal,mackowiak2018cereals} and object
detection~\cite{roy2018deep}. Siddiqui \etal~\cite{viewal} demonstrate that
incorporating multi-view geometry can improve the effectiveness of AL for
semantic segmentation. Yet, the pixel classification formulation in semantic
segmentation makes it easier to adapt AL approaches designed for classification,
while for the keypoint localization task, multi-view adaptations are less
straightforward.

For pose estimation, Yoo~\etal~\cite{yoo2019learning} applied task-agnostic loss
prediction as an AL strategy but with marginal gains over random sampling. Liu
and Ferrari~\cite{liu2017active} propose the Multi-Peak Entropy metric to guide
the sampling of single-view images for annotation. As we demonstrate later
however, extending this metric to multi-view is a non-trivial task.

More recently, Caramalau \etal~\cite{caramalau2021active} extend the
CoreSet~\cite{sener2018active} AL algorithm to hand pose estimation with a
Bayesian formulation. While we also propose an extension to CoreSet in this
paper, our AL strategy relies on geometric intuitions and does not require
expensive Bayesian inference. Additionally, \cite{caramalau2021active} estimates
3D pose from a single depth camera, while we take RGB images from multiple
calibrated cameras as input.

\noindent\textbf{Self-Training and Pseudo-Labeling}: 
Besides active learning,
{self-training}~\cite{xie2020self,zoph2020rethinking,pham2021meta,bengar2021reducing}
is another prominent approach to increasing annotation efficiency. Building on
the principle of knowledge distillation~\cite{distillation}, these methods
perform iterative pseudo-labeling and re-training with unlabeled data. For image
classification, this paradigm has been shown to improve model generalization and
robustness without increasing the amount of human-annotated labels.

For the keypoint localization task in pose estimation, similar ideas have been
explored in the form of semi-supervised learning and
pseudo-labeling~\cite{moskvyak2021semisupervised,Cao_2019_ICCV,honari2018}. In
this paper, inspired by the seminal work of multi-view
bootstrapping~\cite{simon2017hand}, we also develop a pseudo-labeling method.
When applied in conjunction with our AL framework, it leads to even greater
gains in efficiency.


\section{Methods}

The overview of our proposed Active Learning with Self-training system is shown
in Fig.~\ref{fig-overall}. The whole iterative system consists of two main
branches: the active learning branch which selects the unlabeled frames for
human annotation and the self-training branch for pseudo-labeling on the
unlabeled frames. In this section, we first formally define the multi-view pose
estimation problem we are addressing(Sec.~\ref{sec:poseestimationproblem}).
Next, we extend prior works on AL for single-view pose estimation
(Sec.~\ref{sec:extending:AL}). Then, two effective strategies that make full use
of multi-view geometry in the multi-view setting are proposed
(Sec.~\ref{sec:AL}). Additionally, by incorporating pseudo-labels in the
proposed self-training process (Sec.~\ref{sec:SAL}), we show further improvement
in annotation efficiency without extra annotation or computational cost.

\subsection{Pose Estimation Problem Formulation}
\label{sec:poseestimationproblem}
We assume a multi-view capture setup with $\numviews$ synchronized and
calibrated cameras, and we use the term \emph{frame} $\frames$ to denote the
collection of images from all cameras  (\emph{views}) $\view$ at a particular
time instance $t$, \ie $\frames(t) = \left\{\view_1(t), \view_2(t), \cdots,
\view_\numviews(t) \right\}$. In the following, we drop $t$ from the notation
unless necessary. The entire dataset, which is a set of frames (possibly
infinite), is denoted as $\fulldata = \left\{F(1), F(2),\cdots\right\}$.

The task of 3D pose estimation is to estimate the 3D locations of a set of
keypoints on the human body/hand from an input frame. In this work, we focus on
a well-established approach, where the 3D keypoints are obtained by
triangulating 2D predictions on each camera view, using robust triangulation
techniques~\cite{hartley2003multiple}, \eg RANSAC. In particular, the 2D
keypoint prediction problem is formulated as heatmap regression, where  the
ground truth heatmaps are commonly constructed by placing a 2D isotropic
Gaussian at the ground truth location. We use $\totalkp$ to denote the number of
keypoints. 

Note that unlike entropy based AL methods~\cite{liu2017active}, our AL and
self-training system does not limit the pose estimation model to predict a
heatmap for 2D keypoints. Instead, any pose estimation model that performs 2D
keypoint localization and then triangulation would be sufficient.

\subsection{Extending Single-View AL for Pose Estimation}
\label{sec:extending:AL}
Active Learning starts with an initial labeled set $\labeled_0$, and trains an
initial pose estimator. Afterwards, in each iteration $i\geq 1$, an AL strategy
samples a set of frames from the remaining unlabeled set $\unlabeled_i$
following an AL metric $\metric$, queries human annotators, and obtains labels
for them. This enlarges the labeled set $\labeled_i$ into $\labeled_{i+1}$, with
which the pose estimation model is re-trained. Note that $\forall i, \labeled_i
\cup \unlabeled_i = \fulldata$, and that $\labeled_1\subset\labeled_2 \dots
\subset \fulldata$. 

An intuitive approach to AL is to sample examples that receive the most
\emph{uncertain} predictions, and the definition of uncertainty is usually
problem-dependent. The BSB and MPE strategies introduced by Liu
\etal~\cite{liu2017active} fall into this category. 

To our knowledge, no prior work has applied AL to \emph{multi-view} pose
estimation, and the closest work is Liu and Ferrari~\cite{liu2017active}, who
focused on the single-view case.  We thus extend BSB and MPE to multi-view, and
use them as baselines. We extend these single-view strategies by aggregating the
per-view uncertainty metrics, without taking geometry into consideration. In
particular, we focus on the {average}\footnote{We also experimented with other
aggregation functions such as variance, and found them to perform worse.}: if
the per-view predictions have higher uncertainty on average, then,
heuristically, the frame will have higher uncertainty. We define the metric for
the aforementioned entropy-based metrics as
\begin{align}
  \bsb(\frames) & = \frac{1}{\numviews} \sum_{\view \in \frames} \bsb(\view),\\
  \mpe(\frames) & = \frac{1}{\numviews} \sum_{\view \in \frames} \mpe(\view),
\end{align}
where $\bsb(\view)$ and $\mpe(\view)$ are the per-view metrics introduced by Liu
\etal~\cite{liu2017active}. A visualization of these metrics are shown in the
supplementary material.



\subsection{Multi-View AL for Pose Estimation}
\label{sec:AL}

We now discuss AL strategies under the multi-view setting. However, beyond
simple aggregation, the multi-view setting provides extra information to define
geometrically-inspired AL strategies. Recall that the 3D prediction for any
keypoint $\kp$, denoted as $\xyzp^k$, is obtained through robust triangulation;
we will build on this fact to define novel AL strategies. Below, we propose two
AL strategies: \emph{CoreSet-Poses} which is based on pose diversity, and
\emph{Multi-View Consistency} which is based on 3D prediction uncertainty.


\smallskip\noindent\textbf{CoreSet-Poses}: 
CoreSet~\cite{sener2018active} is a state-of-the-art AL strategy based on
selecting diverse representative examples from the unlabeled set, formulated as
solving a combinatorial set-cover problem. Critical to the effectiveness of
CoreSet is modeling the distance between unlabeled examples; in the case of
image classification, Sener \etal~\cite{sener2018active} uses the
Euclidean distance between pretrained convolutional features.
Caramalau \etal~\cite{caramalau2021active} introduced an CoreSet based AL
strategy that only applies to Bayesian pose estimation models. Unlike this prior
work, our proposed CoreSet-Poses strategy can be used on any pose estimation
models.

Our first strategy, {CoreSet-Poses}, builds on CoreSet by supplying it with a
distance metric tailored for pose estimation. Specifically, given a pair of
frames $(\frames,\frames')$, we define their distance $\Delta$ to be the average
Euclidean distance between 3D keypoint predictions
$(\xyzp_{\frames},\xyzp_{\frames'})$ with the current model. While more
sophisticated distance metrics could be defined with respect to the underlying
sets of 2D heatmap predictions, the 3D predictions have already been filtered
through robust triangulation, and have much lower dimensions so distance
computation can be efficient. In practice, we align $\xyzp$ by shifting the root
keypoint to the origin, \eg if the root keypoint is $0$, the aligned pose would
be: $\axyzp = \xyzp - \xyzp^0.$

\begin{algorithm}[t]
\caption{AL for multi-view pose estimation}\label{alg-multi-view}
Input: Labeled set $\labeled$, unlabeled set $\unlabeled$, 
AL metric $\mathcal{M}$,
annotation budget $\budget$;

Sampled Data $S \gets \{\}$\;

\For{$\frames \in \unlabeled$}{
$\mathcal{H}_\frames=\{H_V|\forall V\in \frames\} \gets$ Model Inference;

  $\xyzp_\frames,\terror_\frames\gets \text{triangulate}(\mathcal{H}_\frames)$;
}
\Repeat{$|S| = B$}{
  %

$\frames_\text{greedy} \gets \underset{\frames\in \unlabeled}{\arg\max}\ \mathcal{M}(\frames)$;
\Comment{$\coreset$, $\mc$, \etc}

  $S \gets S \cup \{\frames_\text{greedy}\}$;

  $\labeled \gets \labeled \cup \{\frames_\text{greedy}\}$;

  $\unlabeled \gets \unlabeled \setminus \{\frames_\text{greedy}\}$;
}
\Return{$S$}
\end{algorithm}

Given the distance metric, CoreSet-Poses solves a set-cover problem in order to
maximize coverage in the pose space. While this problem is NP-hard, prior
works~\cite{sener2018active,caramalau2021active} show that it can be
approximately solved by a greedy $k$-center algorithm. Specifically, for each
candidate unlabeled frame $F\in\unlabeled$, we define the CoreSet-Poses AL
metric as
\begin{equation}
  \coreset(F) = \underset{\frames'\in \labeled}{\min} \Delta (\axyzp_\frames, \axyzp_{\frames'}),
\end{equation}
which measures how ``close'' $F$ is to the current labeled set. Then, the greedy
algorithm samples frames with the largest $\coreset$ values. Despite the
improved efficiency, CoreSet-Poses would still take $O(|\unlabeled|^2)$ time to
compute the pairwise distances, making it potentially impractical for large
datasets.

\begin{algorithm}[t]
    \caption{AL + self-training w/ pseudo-labels}\label{alg-self-training}
    Input: Unlabeled set $\unlabeled$, previous pseudo-label set $\pseudo$,
    target amount $\pamount$;
    
    Output: New pseudo-label set $\pseudo'$;

$\pseudo' \gets \{\}, ~\unlabeled' \gets \unlabeled$; \Comment{Make a copy of $\unlabeled$.}

\Repeat(){$|\pseudo'| = \pamount$ \textsc{or} $|\unlabeled| = 0$}{
  $\frames_\text{min} \gets \underset{\frames \in \unlabeled \setminus (\pseudo \cup \pseudo')}{\arg\min}\terror_\frames$; \Comment{No re-labeling.}

  $\unlabeled \gets \unlabeled \setminus \{\frames_\text{min}\}$;

  \If(\Comment{All views are inliers.}){$\numin_{\frames_\text{min}} = \numviews$}{$\pseudo' \gets \pseudo' \cup \{\frames_\text{min}\}$;}
}
$\unlabeled = \unlabeled' \setminus \pseudo'$;

\Return{$\pseudo'$}
\end{algorithm}

\smallskip\noindent\textbf{Multi-view Consistency}:
We now present an uncertainty measure that is intrinsic to the 3D pose
predictions. Our reasoning is that given a frame with multiple views, it is less
likely for the frame-level prediction to be wrong if the per-view 2D predictions
agree with each other. This agreement is in the geometric sense, \eg for two
views, we say two keypoint predictions exactly agree if their epipolar distance
is 0. The corresponding AL strategy is then to sample frames with the largest
disagreement. Additionally, we would like to compute this in $O(|\unlabeled|)$
time, to make it practical for large datasets. We call this the {Multi-View
Consistency} strategy.

Specifically, we take the triangulation error, or the average Euclidean distance
between the 2D keypoint predictions and the reprojected 3D triangulation, as the
AL metric. Note that, since this is exactly the minimization objective for
triangulation, a high error directly indicates strong disagreements between 2D
predictions. Formally, let the predicted 2D location of the $\kp$-th keypoint in
view $\view$ be $l^\kp_\view$, and its reprojected location from the
triangulated $\xyzp^\kp$ be $\hat{l}^\kp_\view$. The triangulation error metric
can be written as:
\begin{equation}
  \mc(\frames) = \frac{1}{\numviews}\frac{1}{\totalkp}
  \sum_{\view\in\frames}\sum_{\kp=1}^\totalkp \|l^\kp_\view-\hat{l}^\kp_\view\|^2.
  \label{eq-error}
\end{equation}
For simplicity, we use $\terror_\frames$ to denote $\mc(\frames)$.

Alg.~\ref{alg-multi-view} presents a unified view of AL for multi-view pose
estimation, where different sampling strategies are realized by choosing the
corresponding metric $\metric$.

\subsection{Improvement via Self-Training}
\label{sec:SAL}

AL is shown to benefit from the addition of techniques like data augmentation
and semi-supervised learning~\cite{mittal2019parting}. In this work, we want to
explore a novel direction to improve it further. We leverage the fact that our
measure of geometric inconsistency can also help us identify reliable frames
with good \emph{pseudo-labels}, which can be directly injected into the training
set. In fact, this is a form of self-training, which has shown great success
recently for image classification
tasks~\cite{xie2020self,zoph2020rethinking,pham2021meta}. These methods use soft
pseudo-labels assigned to unlabeled frames directly and show that the richness
of predictions (compared to a one-hot encoding) is crucial. In the pose
estimation task, the heatmaps can play a similar role, as was demonstrated by
Zhang~\etal~\cite{zhang2019fast} in their work that distills heatmaps from an
8-stack hourglass model to a 4-stack one. However, this approach is not suitable
to make the best use of multi-view predictions, which is the direction we
explore. To take full advantage of multi-view predictions, we project the 3D
keypoints formed by triangulation back to each camera view, and assign
psuedo-heatmaps to a set of frames with the most inliers and with the smallest
triangulation error (Equation~\ref{eq-error}). These predictions are the most
likely to be closest to the actual ground truth, thus they are excellent
candidates to be used in self-training.

We call this the pseudo-label set $\pseudo$, and we augment the training set to
be $\pseudo \cup \labeled$ in each AL iteration. Similar to multi-view
bootstrapping~\cite{simon2017hand}, our motivation is that by adding $\pseudo$
to the training set, the model is exposed to more varied data and can learn to
generalize better. However, the proposed self-training algorithm is able to
avoid ``model drifting'' in iterative training, using an entirely
\emph{automated} strategy, as shown in Fig.~\ref{fig-sal-dist}.  This is in
contrast to multi-view bootstrapping~\cite{simon2017hand}, which requires human
verification in the loop.

Contrary to AL, self-training requires the pseudo-labels to be confident and
accurate, and careful selection is key. Simon \etal~\cite{simon2017hand} uses
heuristics specific to hand anatomy to filter candidate frames, and conducts
additional human verification. Instead, our approach is fully automated.
Specifically, for a pseudo-labeled frame to be considered for selection, we
require that all views for all keypoints to be {inliers} during triangulation.
Then, we take candidate frames with the smallest triangulation error
$\terror_\frames$, that are \emph{not already selected} in the previous AL
iteration, to form the pseudo-label set $\pseudo$. We found the latter heuristic
to be critical in preventing drifting of the pseudo-labels. Our {self-training}
algorithm is summarized in Alg.~\ref{alg-self-training}.


\section{Experiments}\label{sec:exp}

\begin{figure*}[t]
  \centering
  \begin{subfigure}{0.32\textwidth}
    \resizebox{\columnwidth}{!}{
      \begin{tikzpicture}
        \begin{axis}[
            grid=both,
            grid style={line width=.1pt, draw=gray!40},
            title = Panoptic / PoseResNet-50,
            ylabel= 3D MKPE (mm),
            xlabel= Annotation Cost,
            xticklabels={0,0,$5\%$,$10\%$,$15\%$,$20\%$},
            yticklabels={1.0,2.0,3.0,4.0,5.0,6.0},
            legend pos = north east,
            legend style={font=\small, fill=white, fill opacity=0.5, draw opacity=1, text opacity=1},
            legend cell align={left},
            cycle list name=default,
            scaled x ticks = false,
            ymin=2.3,ymax=5.6,
          ]
          \addplot table[x={PERCENT},y={AL-R-P-RESNET}, y error={AL-R-P-RESNET-ERROR}]{data/p.txt};
          \addplot table[x={PERCENT},y={AL-BSB-P-RESNET}, y error={AL-BSB-P-RESNET-ERROR}]{data/p.txt};
          \addplot table[x={PERCENT},y={AL-MPE-P-RESNET}, y error={AL-MPE-P-RESNET-ERROR}]{data/p.txt};
          \addplot table[x={PERCENT},y={AL-CS-P-RESNET}, y error={AL-CS-P-RESNET-ERROR}]{data/p.txt};
          \addplot table[x={PERCENT},y={AL-T-P-RESNET}, y error={AL-T-P-RESNET-ERROR}]{data/p.txt};
        \end{axis}
      \end{tikzpicture}
    }
  \end{subfigure}
  \begin{subfigure}{0.32\textwidth}
    \resizebox{\columnwidth}{!}{
      \begin{tikzpicture}
        \begin{axis}[
            grid=both,
            grid style={line width=.1pt, draw=gray!40},
            title = Panoptic / HRNet,
            ylabel= 3D MKPE (mm),
            xlabel= Annotation Cost,
            xticklabels={0,0,$5\%$,$10\%$,$15\%$,$20\%$},
            legend pos = north east,
            legend style={font=\small, fill=white, fill opacity=0.5, draw opacity=1, text opacity=1},
            legend cell align={left},
            cycle list name=default,
            scaled x ticks = false,
            ymin=2.2, ymax=4.3
          ]
          \addplot table[x={PERCENT},y={AL-R-P-HR}, y error={AL-R-P-HR-ERROR}]{data/p.txt};
          \addplot table[x={PERCENT},y={AL-BSB-P-HR}, y error={AL-BSB-P-HR-ERROR}]{data/p.txt};
          \addplot table[x={PERCENT},y={AL-MPE-P-HR}, y error={AL-MPE-P-HR-ERROR}]{data/p.txt};
          \addplot table[x={PERCENT},y={AL-CS-P-HR}, y error={AL-CS-P-HR-ERROR}]{data/p.txt};
          \addplot table[x={PERCENT},y={AL-T-P-HR}, y error={AL-T-P-HR-ERROR}]{data/p.txt};
        \end{axis}
      \end{tikzpicture}
    }
  \end{subfigure}
  \begin{subfigure}{0.32\textwidth}
    \resizebox{\columnwidth}{!}{
      \begin{tikzpicture}
        \begin{axis}[
            grid=both,
            grid style={line width=.1pt, draw=gray!40},
            title = InterHand / PoseResNet-50,
            ylabel= 3D MKPE (mm),
            xlabel= Annotation Cost,
        xtick={20,30,40,50},
        xticklabels={20$\%$,30$\%$,40$\%$,50$\%$},
            legend pos = north east,
            legend style={font=\small, fill=white, fill opacity=0.5, draw opacity=1, text opacity=1},
            legend cell align={left},
            cycle list name=default,
            scaled x ticks = false,
            ymin=2.2,ymax=4.3
          ]
          \addplot table[x={PERCENT},y={AL-R-IH-RESNET}, y error={AL-R-IH-RESNET-ERROR}]{data/ih.txt};
          \addplot table[x={PERCENT},y={AL-BSB-IH-RESNET}, y error={AL-BSB-IH-RESNET-ERROR},col sep=comma]{data/interhand-1107.txt};
          \addplot table[x={PERCENT},y={AL-MPE-IH-RESNET}, y error={AL-MPE-IH-RESNET-ERROR},col sep=comma]{data/interhand-1107.txt};
          \addplot table[x={PERCENT},y={AL-CORESET-IH-RESNET}, y error={AL-CORESET-IH-RESNET-ERROR},col sep=comma]{data/interhand-1107.txt};
          \addplot table[x={PERCENT},y={AL-T-IH-RESNET}, y error={AL-T-IH-RESNET-ERROR}]{data/ih.txt};
        \end{axis}
      \end{tikzpicture}
    }
  \end{subfigure}
  \begin{subfigure}{\textwidth}
    \centering
    \resizebox{!}{1.4em}{
      \begin{tikzpicture}
        \begin{customlegend}[anchor=north west, legend columns=6,
            legend style={fill=none,draw=none,align=center,column sep=2ex},
            legend entries={\random,
              \BSB,
              \MPE,
              \CS,
              \MC,
          }]
          \addlegendimage{mark=x, color=red}
          \addlegendimage{mark=x, color=blue}
          \addlegendimage{mark=x, color=cyan}
          \addlegendimage{mark=x, color=violet}
          \addlegendimage{mark=x, color=C3}
        \end{customlegend}
      \end{tikzpicture}
    }
  \end{subfigure}
  \vspace{-1em}
  \caption{
    Comparison of AL strategies on \panoptic{} and \ih{}. X-axis: percent of
    dataset labeled. 
    BSB and MPE~\cite{liu2017active}, developed for single-view pose estimation,
    do not perform better than \random{} when extended to multi-view. Our
    proposed strategies (\CS{} and \MC{}) significantly outperform random
    sampling. Best viewed in color.
  }
  \label{fig-al-p}
  \vspace{-0.1in}
\end{figure*}
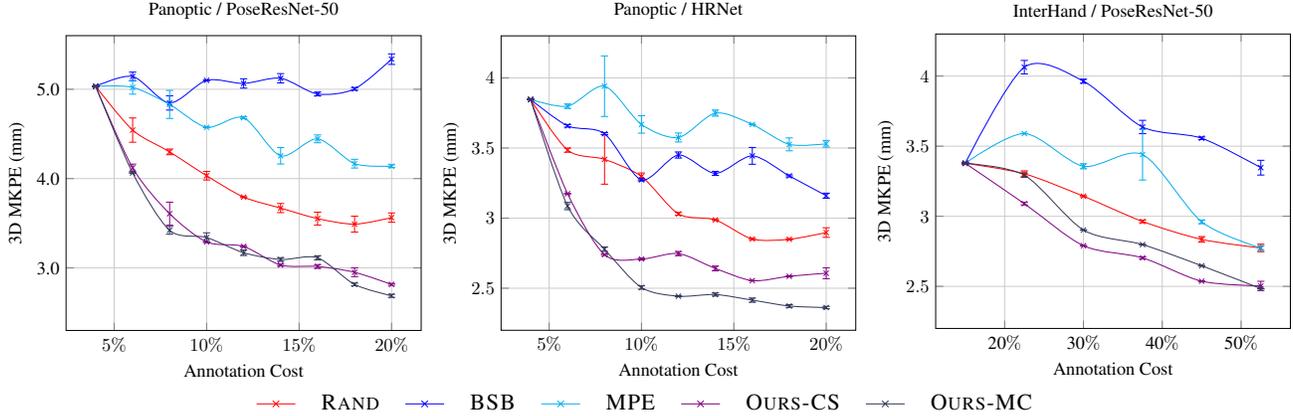

\subsection{Datasets and Evaluation}
\label{sec:exp:setup}
To simulate the data annotation process, we use two large-scale multi-view
benchmarks in our experiments: \panoptic~\cite{joo2017panoptic} for the body
pose estimation problem, and \ih~\cite{moon2020interhand} for the hand pose
estimation problem.

The \panoptic{} dataset has 9 sequences each having 31 camera views, and over
160,000 frames in total.  We split them into 7 sequences for training, 1
sequence for validation and 1 sequence for test.  We use 8 eye-level cameras for
training and validation and 30 cameras\footnote{Video from 1 test camera is
missing on CMU website.} for testing, including the 8 eye-level cameras used
during training and validation. Sequences are temporally sub-sampled at 1 frame
per second, and we end up with 5,008 training frames (40,064 images), 891
validation frames (7,128 images) and 771 test frames (23,130 images).
We use the 5fps version of \ih, and sub-sample the dataset into 10 Captures for
training, 1 Capture for validation and another 1 Capture for testing.  For each
capture, we use 16 cameras that are distantly-located during training and
validation. Moreover, we use 32 cameras during testing. We end up with 12,123
training frames (193,968 images), 1,900 validation frames (30,400 images) and
1,762 test frames (56,384 images).

For each experiment, we conduct 3 randomized trials, and report the average and
variance for the 3D Mean Key Point Error (MKPE) in millimeter (mm). As our
backbone models predict 2D heatmaps for each view, to obtain the 3D prediction
$\xyzp^k$ we perform RANSAC triangulation with the 2D keypoint predictions
$\xyp^\kp$ (argmax of the heatmap).

\subsection{Annotation Simulation Details}
\label{sec:exp:implementation}
We use two backbone models in our experiments:
PoseResNet-50~\cite{xiao2018simple} and HRNet~\cite{wang2020deep}. For body pose
estimation, both backbones are pretrained on the MPII~\cite{andriluka20142d}
dataset. As MPII and \panoptic{} define different sets of keypoints, we
initialize the weights of all layers except the output layer of the
PoseResNet-50. For HRNet, we use the pretrained weights of the first 4 layers
and randomly initialize the remaining layers. For hand pose estimation,
as no pretrained models for our setting are available, we randomly initialize
all parameters from a normal distribution.

The annotation amount in each AL iteration is set to 100 frames for \panoptic,
and 1,000 frames for \ih. Regardless of the AL strategy, frames in the initial
labeled set $\labeled_0$ (200 frames for \panoptic, 2000 frames for \ih) are
always randomly sampled, to provide a reasonable starting point. Furthermore,
for the sake of reproducibility, all strategies start with the same set of
randomly sampled frames. For self-training, the pseudo-label amount is set to
10\%-20\% of the annotation amount.

In each AL iteration, we train the model from scratch on the current labeled
dataset, as well as the pseudo-labeled dataset if available. Both backbones are
trained with a batch size of 32 images per GPU for a total of 5000 optimization
steps. We use Adam optimizer with a learning rate starting at 0.001, and decayed
by 1/10 at the midpoint. All experiments are carried out with this training
procedure, and all reported results are evaluated on the same held-out set.

Following Mittal \etal~\cite{mittal2019parting}, we also
experiment with data augmentation. We use
RandAugment~\cite{cubuk2020randaugment} to augment the training images for
\panoptic. On the other hand, RandAugment does not result in better performances
on \ih, which contains more diverse poses.

\begin{figure*}[t]
  \centering
  \begin{subfigure}{0.32\textwidth}
    \centering
    \resizebox{\columnwidth}{!}{
      \begin{tikzpicture}
        \begin{axis}[
            grid=both,
            grid style={line width=.1pt, draw=gray!40},
            title = Panoptic / PoseResNet-50 / RandAug,
            ylabel= 3D MKPE (mm),
            xlabel= Annotation Cost,
            xtick={5, 10, 15},
            xticklabels={$5\%$,$10\%$,$15\%$},
            legend pos = north east,
            legend style={font=\small, fill=white, fill opacity=0.6, draw opacity=1, text opacity=1},
            legend cell align={left},
            cycle list name=default,
            scaled x ticks = false,
            ymin=2.2,ymax=4.7
          ]
          \addplot table[restrict x to domain=1:16, x={PERCENT}, y={AUG-R-P-RESNET}, y error={AUG-R-P-RESNET-ERROR}] {data/p.txt};
          \addplot table[restrict x to domain=1:16, x={PERCENT}, y={SAL-25-R-P-RESNET}, y error={SAL-25-R-P-RESNET-ERROR}] {data/p.txt};
          \addplot table[restrict x to domain=1:16, x={PERCENT}, y={AUG-T-P-RESNET}, y error={AUG-T-P-RESNET-ERROR}] {data/p.txt};
          \addplot table[restrict x to domain=1:16, x={PERCENT}, y={SAL-25-T-P-RESNET}, y error={SAL-25-T-P-RESNET-ERROR}] {data/p.txt};
        \end{axis}
      \end{tikzpicture}
    } 
  \end{subfigure}
  \begin{subfigure}{0.32\textwidth}
    \centering
    \resizebox{\columnwidth}{!}{
      \begin{tikzpicture}
        \begin{axis}[
            grid=both,
            grid style={line width=.1pt, draw=gray!40},
            title = Panoptic / HRNet / RandAug,
            ylabel= 3D MKPE (mm),
            xlabel= Annotation Cost,
            xtick={5, 10, 15},
            xticklabels={$5\%$,$10\%$,$15\%$},
            legend pos = north east,
            legend style={font=\small, fill=white, fill opacity=0.6, draw opacity=1, text opacity=1},
            legend cell align={left},
            cycle list name=default,
            scaled x ticks = false,
            ymin=2.2,ymax=3.7
          ]
          \addplot table[restrict x to domain=1:16, x={PERCENT},y={AUG-R-P-HR}, y error={AUG-R-P-HR-ERROR}, col sep=comma]{data/panoptic-1106.txt}; 
          \addplot table[restrict x to domain=1:16, x={PERCENT},y={SAL-25-R-P-HR}, y error={SAL-25-R-P-HR-ERROR}]{data/p.txt}; 
          \addplot table[restrict x to domain=1:16, x={PERCENT},y={AUG-T-P-HR}, y error={AUG-T-P-HR-ERROR}]{data/p.txt};
          \addplot table[restrict x to domain=1:16, x={PERCENT},y={SAL-25-T-P-HR}, y error={SAL-25-T-P-HR-ERROR}]{data/p.txt};
        \end{axis}
      \end{tikzpicture}
    } 
  \end{subfigure}
  \begin{subfigure}{0.32\textwidth}
    \centering
    \resizebox{\columnwidth}{!}{
      \begin{tikzpicture}
        \begin{axis}[
            grid=both,
            grid style={line width=.1pt, draw=gray!40},
            title = InterHand / PoseResNet-50,
            ylabel= 3D MKPE (mm),
            xlabel= Annotation Cost,
            xtick={20,30,40,50},
            xticklabels={20$\%$,30$\%$,40$\%$,50$\%$},
            legend pos = north east,
            legend style={font=\small, fill=white, fill opacity=0.6, draw opacity=1, text opacity=1},
            legend cell align={left},
            cycle list name=default,
            scaled x ticks = false,
            ymin=2.2,ymax=3.7
          ]
          \addplot table[x={PERCENT},y={AL-R-IH-RESNET}, y error={AL-R-IH-RESNET-ERROR}]{data/ih.txt};
          \addplot table[x={PERCENT},y={SALNOAUG-100-R-IH-RESNET}, y error={SALNOAUG-100-R-IH-RESNET-ERROR}]{data/ih.txt};
          \addplot table[x={PERCENT},y={AL-T-IH-RESNET}, y error={AL-T-IH-RESNET-ERROR}]{data/ih.txt};
          \addplot table[x={PERCENT},y={SALNOAUG-100-T-IH-RESNET}, y error={SALNOAUG-100-T-IH-RESNET-ERROR}]{data/ih.txt};
        \end{axis}
      \end{tikzpicture}
    } 
  \end{subfigure}
  \begin{subfigure}{\textwidth}
    \centering
    \resizebox{!}{1.4em}{
      \begin{tikzpicture}
        \begin{customlegend}[anchor=north west, legend columns=6,
            legend style={fill=none,draw=none,align=center,column sep=2ex},
            legend entries={$\random$,
              $\random+\sal$,
              $\MC$,
              $\MC+\sal$,
          }]
          \addlegendimage{mark=x, color=red}
          \addlegendimage{mark=x, color=blue}
          \addlegendimage{mark=x, color=cyan}
          \addlegendimage{mark=x, color=violet}
        \end{customlegend}
      \end{tikzpicture}
    }
  \end{subfigure}
  \vspace{-1em}
  \caption{
    AL + self-training ($\sal$) on \panoptic{}  and \ih{}. X-axis: percent of
    dataset labeled. 
    When combined with AL, our automated self-training strategy enables
    additional label efficiency gains at no extra computational cost, especially
    during the early stages of training. Best viewed in color.
  }
  \label{fig-al-vs-sal}
  \vspace{-0.1in}
\end{figure*}
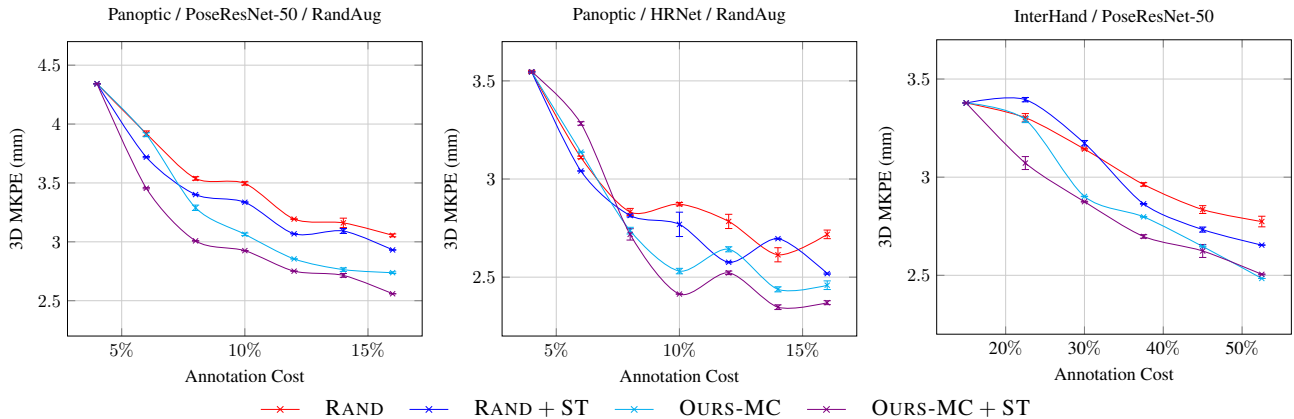
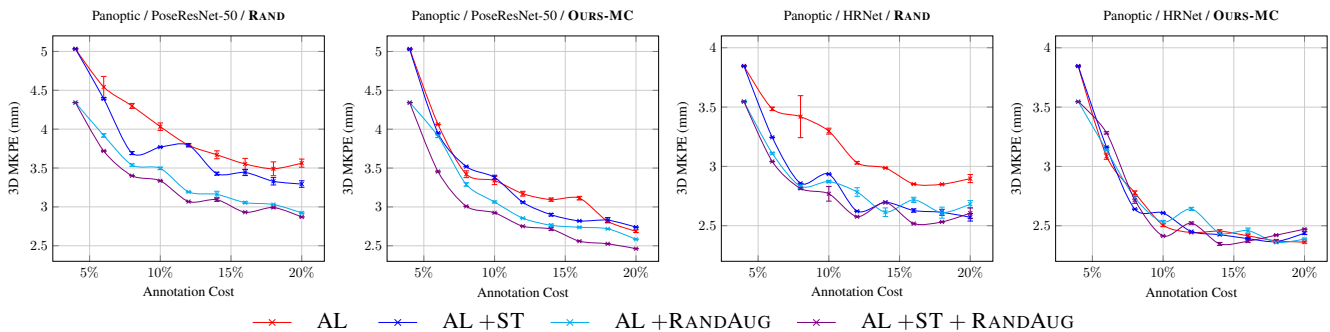
\begin{figure*}[t]
    \centering
    \begin{subfigure}{0.245\textwidth}
      \resizebox{\columnwidth}{!}{
        \begin{tikzpicture}
          \begin{axis}[
              grid=both,
              grid style={line width=.1pt, draw=gray!40},
              title = Panoptic / PoseResNet-50 / \textbf{\random},
              ylabel= 3D MKPE (mm),
              xlabel= Annotation Cost,
              xticklabels={0,0,$5\%$,$10\%$,$15\%$,$20\%$},
              legend pos = north east,
              legend style={font=\small, fill=white, fill opacity=0.5, draw opacity=1, text opacity=1},
              legend cell align={left},
              cycle list name=default,
              scaled x ticks = false,
              ymin=2.3,ymax=5.2,
            ]
            \addplot table[x={PERCENT},y={AL-R-P-RESNET}, y error={AL-R-P-RESNET-ERROR}]{data/p.txt};
            \addplot table[x={PERCENT},y={SALNOAUG-10-R-P-RESNET}, y error={SALNOAUG-10-R-P-RESNET-ERROR}]{data/p.txt};
            \addplot table[x={PERCENT},y={AUG-R-P-RESNET}, y error={AUG-R-P-RESNET-ERROR}]{data/p.txt};
            \addplot table[x={PERCENT}, y={SAL-25-R-P-RESNET}, y error={SAL-25-R-P-RESNET-ERROR}] {data/p.txt};
          \end{axis}
        \end{tikzpicture}
      }
    \end{subfigure}
    \begin{subfigure}{0.245\textwidth}
      \resizebox{\columnwidth}{!}{
        \begin{tikzpicture}
          \begin{axis}[
              grid=both,
              grid style={line width=.1pt, draw=gray!40},
              title = Panoptic / PoseResNet-50 / \textbf{\MC},
              ylabel= 3D MKPE (mm),
              xlabel= Annotation Cost,
              xticklabels={0,0,$5\%$,$10\%$,$15\%$,$20\%$},
              legend pos = north east,
              legend style={font=\small, fill=white, fill opacity=0.5, draw opacity=1, text opacity=1},
              legend cell align={left},
              cycle list name=default,
              scaled x ticks = false,
              ymin=2.3,ymax=5.2,
            ]
            \addplot table[x={PERCENT},y={AL-T-P-RESNET}, y error={AL-T-P-RESNET-ERROR}]{data/p.txt};
            \addplot table[x={PERCENT},y={SALNOAUG-10-T-P-RESNET}, y error={SALNOAUG-10-T-P-RESNET-ERROR}]{data/sal-panoptics-triangulation.txt};
            \addplot table[x={PERCENT},y={AUG-T-P-RESNET}, y error={AUG-T-P-RESNET-ERROR}]{data/p.txt};
            \addplot table[x={PERCENT}, y={SAL-25-T-P-RESNET}, y error={SAL-25-T-P-RESNET-ERROR}] {data/p.txt};
          \end{axis}
        \end{tikzpicture}
      }
    \end{subfigure}
    \begin{subfigure}{0.245\textwidth}
        \resizebox{\columnwidth}{!}{
          \begin{tikzpicture}
            \begin{axis}[
                grid=both,
                grid style={line width=.1pt, draw=gray!40},
                title = Panoptic / HRNet / \textbf{\random},
                ylabel= 3D MKPE (mm),
                xlabel= Annotation Cost,
                xticklabels={0,0,$5\%$,$10\%$,$15\%$,$20\%$},
                legend pos = north east,
                legend style={font=\small, fill=white, fill opacity=0.5, draw opacity=1, text opacity=1},
                legend cell align={left},
                cycle list name=default,
                scaled x ticks = false,
                ymin=2.2,ymax=4.1,
              ]
              \addplot table[x={PERCENT},y={AL-R-P-HR}, y error={AL-R-P-HR-ERROR}]{data/p.txt};
              \addplot table[x={PERCENT},y={SALNOAUG-10-R-P-HR}, y error={SALNOAUG-10-R-P-HR-ERROR}]{data/p.txt};
              \addplot table[x={PERCENT},y={AUG-R-P-HR}, y error={AUG-R-P-HR-ERROR}, col sep=comma]{data/panoptic-1106.txt}; 
              \addplot table[x={PERCENT}, y={SAL-25-R-P-HR}, y error={SAL-25-R-P-HR-ERROR}] {data/p.txt};
            \end{axis}
          \end{tikzpicture}
        }
      \end{subfigure}
      \begin{subfigure}{0.245\textwidth}
        \resizebox{\columnwidth}{!}{
          \begin{tikzpicture}
            \begin{axis}[
                grid=both,
                grid style={line width=.1pt, draw=gray!40},
                title = Panoptic / HRNet / \textbf{\MC},
                ylabel= 3D MKPE (mm),
                xlabel= Annotation Cost,
                xticklabels={0,0,$5\%$,$10\%$,$15\%$,$20\%$},
                legend pos = north east,
                legend style={font=\small, fill=white, fill opacity=0.5, draw opacity=1, text opacity=1},
                legend cell align={left},
                cycle list name=default,
                scaled x ticks = false,
                ymin=2.2,ymax=4.1,
              ]
              \addplot table[x={PERCENT},y={AL-T-P-HR}, y error={AL-T-P-HR-ERROR}]{data/p.txt};
              \addplot table[x={PERCENT},y={SALNOAUG-10-T-P-HR}, y error={SALNOAUG-10-T-P-HR-ERROR}]{data/p.txt};
              \addplot table[x={PERCENT},y={AUG-T-P-HR}, y error={AUG-T-P-HR-ERROR}]{data/p.txt};
              \addplot table[x={PERCENT}, y={SAL-25-T-P-HR}, y error={SAL-25-T-P-HR-ERROR}] {data/p.txt};
            \end{axis}
          \end{tikzpicture}
        }
      \end{subfigure}
    \begin{subfigure}{\textwidth}
      \centering
      \resizebox{!}{1.4em}{
        \begin{tikzpicture}
          \begin{customlegend}[anchor=north west, legend columns=6,
              legend style={fill=none,draw=none,align=center,column sep=2ex},
              legend entries={AL ,
                AL $+\sal$,
                AL $+\textsc{RandAug}$,
                AL $+\sal+\textsc{RandAug}$,
            }]
            \addlegendimage{mark=x, color=red}
            \addlegendimage{mark=x, color=blue}
            \addlegendimage{mark=x, color=cyan}
            \addlegendimage{mark=x, color=violet}
          \end{customlegend}
        \end{tikzpicture}
      }
    \end{subfigure}
    \caption{ 
      Comparison between AL, AL $+$ self-training ($\sal$), AL
      $+\textsc{RandAug}$, and AL $+\sal+\textsc{RandAug}$ on \panoptic{}.
      X-axis: percent of dataset labeled. 
      Our self-training strategy provides a large improvement on PoseResNet with
      the \random{} AL strategy. Although RandAug further improves the
      generalization of the PoseResNet and HRNet for both \random{} and \MC{} AL
      strategies, our self-training strategy still shows a minor improvement
      from the AL only baseline.
    }
    \label{fig-overall-comparison}
    \vspace{-0.1in}
  \end{figure*}
  

\subsection{Results}
We experiment with both PoseResNet-50 and HRNet on \panoptic, while for the much
larger \ih{} we report results from PoseResNet-50. Below, we refer to random
sampling as $\random$, Multi-Peak Entropy strategy~\cite{liu2017active} as
$\MPE$, Best vs. Second Best strategy~\cite{liu2017active} as $\BSB$, our
proposed CoreSet-Poses strategy as $\CS$, and Multi-View Consistency strategy as
$\MC$. 


\subsubsection{Active Learning}
Results with PoseResNet-50 and HRNet on \panoptic{} and PoseResNet-50 on \ih{}
are reported in Fig.~\ref{fig-al-p}. We do not use data augmentation in this
experiment in order to highlight the differences in sampling strategies.

As we mentioned earlier, the \random{} strategy can be a very strong baseline
for difficult tasks like pose estimation. Although MPE has been reported to
outperform \random{} in single-view pose estimation~\cite{liu2017active},  we
observe that extending $\MPE$ or $\BSB$ to multi-view by aggregating per-frame
uncertainty measures fails to beat \random. Furthermore, simple forms of
aggregation also fail to account for the geometric structure in the problem: it
is possible that all 2D predictions are highly confident, while being
geometrically inconsistent. In such cases, the frame would fail the
triangulation, yet still score low enough with $\MPE$ and $\BSB$ to evade
selection.

Next, our proposed strategies $\MC$ and $\CS$ outperform \random{} consistently
by a large margin in all scenarios. $\MC$ is on par with $\CS$ with the
PoseResNet-50 backbone, but outperforms the $\CS$ with the HRNet backbone,
despite only taking a fraction of the computational cost on the unlabeled set
($O(|\unlabeled|)$ vs. $O(|\unlabeled|^2$)). 

The improvement of $\MC$ compared to $\random$ on \ih{} is smaller than that on
\panoptic, as the variations of poses in \ih{} is much higher than \panoptic.
Additionally, we sample frames from \ih{} more sparsely than \panoptic. As
mentioned above, \ih{} is much larger and contains more diverse poses than
\panoptic, \ie diverse samples can be achieved by random sampling when the
unlabeled set is diverse. Therefore, we conduct our ablation studies mainly on
\panoptic.


\subsubsection{AL + Self-Training}
For this experiment, we focus on building a complete system: we use
pseudo-labels to augment the training set in AL iterations, and we add data
augmentation (except for \ih{} as previously mentioned). For clarity, we pick
the overall best method from the previous experiment, $\MC$, and compare it
against \random{}. Results are shown in Fig~\ref{fig-al-vs-sal}.

Similar to the findings in multi-view bootstrapping~\cite{simon2017hand}, the
additional self-training process provides consistent improvements to active
learning. In our problem setting, we also observe the benefits to be more
pronounced at the early stages: for example, on \panoptic{} with 10\% data
annotated, pseudo-labels reduce the gap between 10\% and 20\% annotated data
amount by 20\% with the PoseResNet-50 backbone, and by around $50\%$ for HRNet.

We find that pseudo-labels would negatively drift if the pseudo-labeled frames
are sampled from $\unlabeled$ instead of $\unlabeled \setminus \pseudo$ in each
iteration. Essentially, the same frames would keep re-entering $\pseudo$ and
their labels become worse over each AL iteration. The number of frames to
include, $M$, is also a crucial parameter. We present more ablative studies
regarding these design choices in the supplementary material.

In summary, the above results show that our proposed AL strategies outperform
the baselines steadily by a large margin, for both body and hand pose
estimation. Additionally, with a carefully tuned self-training process, we can
further improve annotation efficiency, with no extra cost.

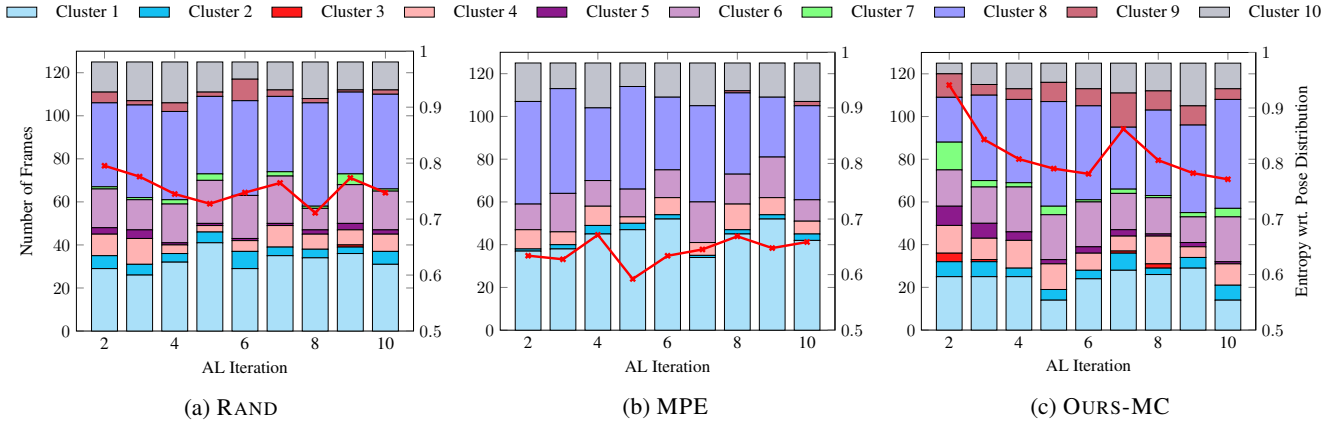
\begin{figure*}
\centering
\begin{subfigure}{\textwidth}
  \centering
  \resizebox{\columnwidth}{!}{
    \begin{tikzpicture}
      \begin{customlegend}[anchor=north west, legend columns=10,
          legend style={draw=none,align=center,column sep=2ex},
          legend entries={
            Cluster 1,
            Cluster 2,
            Cluster 3,
            Cluster 4,
            Cluster 5,
            Cluster 6,
            Cluster 7,
            Cluster 8,
            Cluster 9,
            Cluster 10,
        }]
        \addlegendimage{area legend,fill=cyan!30!white}
        \addlegendimage{area legend,fill=cyan!70!white}
        \addlegendimage{area legend,fill=red!90!white}
        \addlegendimage{area legend,fill=red!30!white}
        \addlegendimage{area legend,fill=violet!90!white}
        \addlegendimage{area legend,fill=violet!40!white}
        \addlegendimage{area legend,fill=green!50!white}
        \addlegendimage{area legend,fill=blue!40!white}
        \addlegendimage{area legend,fill=C2!90!white}
        \addlegendimage{area legend,fill=C3!30!white}
      \end{customlegend}
    \end{tikzpicture}
  }
\end{subfigure}
\vspace{-.5em}

\begin{subfigure}{0.33\textwidth}
  \resizebox{\columnwidth}{!}{
    \begin{tikzpicture}
      \begin{axis}[
          ybar stacked,
          bar width=15pt,        
          xlabel={AL Iteration},
          ylabel={Number of Frames},
          symbolic x coords={2,3,4,5,6,7,8,9,10},
          cycle list name=cluster,
          ymin=0, ymax=130
        ]
        \addplot+[ybar] plot coordinates {(2,29)(3,26)(4,32)(5,41)(6,29)(7,35)(8,34)(9,36)(10,31)};
        \addplot+[ybar] plot coordinates {(2,6)(3,5)(4,4)(5,5)(6,8)(7,4)(8,4)(9,3)(10,6)};        
        \addplot+[ybar] plot coordinates {(2,0)(3,0)(4,0)(5,0)(6,0)(7,0)(8,0)(9,1)(10,0)};
        \addplot+[ybar] plot coordinates {(2,10)(3,12)(4,4)(5,3)(6,5)(7,10)(8,7)(9,7)(10,8)};
        \addplot+[ybar] plot coordinates {(2,3)(3,4)(4,1)(5,1)(6,1)(7,1)(8,2)(9,3)(10,2)};
        \addplot+[ybar] plot coordinates {(2,18)(3,14)(4,18)(5,20)(6,20)(7,22)(8,10)(9,18)(10,18)};
        \addplot+[ybar] plot coordinates {(2,1)(3,1)(4,2)(5,3)(6,0)(7,2)(8,1)(9,5)(10,1)};        
        \addplot+[ybar] plot coordinates {(2,39)(3,43)(4,41)(5,36)(6,44)(7,35)(8,48)(9,38)(10,44)};
        \addplot+[ybar] plot coordinates {(2,5)(3,2)(4,4)(5,2)(6,10)(7,3)(8,2)(9,1)(10,2)};
        \addplot+[ybar] plot coordinates {(2,14)(3,18)(4,19)(5,14)(6,8)(7,13)(8,17)(9,13)(10,13)};
      \end{axis}
      \begin{axis}[
          axis y line*=right,
          axis x line=none,
          ymin=0.5, ymax=1,
        ]
        \addplot [mark=x, color=red, line width=0.5mm] coordinates
        {(2,0.795334969067265)(3,0.775914227732779)(4,0.744854486376273)(5,0.727493548150196)(6,0.747419217557714)(7,0.764582097940892)(8,0.711185689288352)(9,0.773645219399997)(10,0.747165642392085)};
      \end{axis}
    \end{tikzpicture}
  }
  \caption{\random}
\end{subfigure}
\begin{subfigure}{0.31\textwidth}
  \resizebox{\columnwidth}{!}{
    \begin{tikzpicture}
      \begin{axis}[
          ybar stacked,
          bar width=15pt,
          legend pos=outer north east,
          xlabel={AL Iteration},
          symbolic x coords={2,3,4,5,6,7,8,9,10},
          cycle list name=cluster,
          reverse legend,
          ymin=0, ymax=130,
        ]
        \addplot+[ybar] plot coordinates {(2,37)(3,38)(4,45)(5,47)(6,52)(7,34)(8,45)(9,52)(10,42)};
        \addplot+[ybar] plot coordinates {(2,1)(3,2)(4,4)(5,3)(6,2)(7,1)(8,2)(9,2)(10,3)};
        \addplot+[ybar] plot coordinates {(2,0)(3,0)(4,0)(5,0)(6,0)(7,0)(8,0)(9,0)(10,0)};
        \addplot+[ybar] plot coordinates {(2,9)(3,6)(4,9)(5,3)(6,8)(7,6)(8,12)(9,8)(10,6)};
        \addplot+[ybar] plot coordinates {(2,0)(3,0)(4,0)(5,0)(6,0)(7,0)(8,0)(9,0)(10,0)};
        \addplot+[ybar] plot coordinates {(2,12)(3,18)(4,12)(5,13)(6,13)(7,19)(8,14)(9,19)(10,10)};
        \addplot+[ybar] plot coordinates {(2,0)(3,0)(4,0)(5,0)(6,0)(7,0)(8,0)(9,0)(10,0)};
        \addplot+[ybar] plot coordinates {(2,48)(3,49)(4,34)(5,48)(6,34)(7,45)(8,38)(9,28)(10,44)};
        \addplot+[ybar] plot coordinates {(2,0)(3,0)(4,0)(5,0)(6,0)(7,0)(8,1)(9,0)(10,2)};
        \addplot+[ybar] plot coordinates {(2,18)(3,12)(4,21)(5,11)(6,16)(7,20)(8,13)(9,16)(10,18)};
      \end{axis}
      \begin{axis}[
          axis y line*=right,
          axis x line=none,
          ymin=0.5, ymax=1,
        ]
        \addplot [mark=x, color=red, line width=0.5mm] coordinates
        {(2,0.634059566496921)(3,0.627570555960882)(4,0.671485621075443)(5,0.592210098111623)(6,0.633898651102509)(7,0.645304635943957)(8,0.668864966490353)(9,0.647777078812426)(10,0.658625018514037)};
      \end{axis}
    \end{tikzpicture}
  }
  \caption{$\MPE$}
\end{subfigure}
\begin{subfigure}{0.33\textwidth}
  \resizebox{\columnwidth}{!}{
    \begin{tikzpicture}
      \begin{axis}[
          ybar stacked,
          bar width=15pt,
          reverse legend,
          legend cell align={left},
          legend style={at={(1.55,1)}},
          xlabel={AL Iteration},
          symbolic x coords={2,3,4,5,6,7,8,9,10},
          cycle list name=cluster,
          ymin=0, ymax=130
        ]
        \addplot+[ybar] plot coordinates {(2,25)(3,25)(4,25)(5,14)(6,24)(7,28)(8,26)(9,29)(10,14)};
        \addplot+[ybar] plot coordinates {(2,7)(3,7)(4,4)(5,5)(6,4)(7,8)(8,3)(9,5)(10,7)};
        \addplot+[ybar] plot coordinates {(2,4)(3,1)(4,0)(5,0)(6,0)(7,1)(8,2)(9,0)(10,0)};
        \addplot+[ybar] plot coordinates {(2,13)(3,10)(4,13)(5,12)(6,8)(7,7)(8,13)(9,5)(10,10)};
        \addplot+[ybar] plot coordinates {(2,9)(3,7)(4,4)(5,2)(6,3)(7,3)(8,1)(9,2)(10,1)};
        \addplot+[ybar] plot coordinates {(2,17)(3,17)(4,21)(5,21)(6,21)(7,17)(8,17)(9,12)(10,21)};
        \addplot+[ybar] plot coordinates {(2,13)(3,3)(4,2)(5,4)(6,1)(7,2)(8,1)(9,2)(10,4)};
        \addplot+[ybar] plot coordinates {(2,21)(3,40)(4,39)(5,49)(6,44)(7,29)(8,40)(9,41)(10,51)};
        \addplot+[ybar] plot coordinates {(2,11)(3,5)(4,5)(5,9)(6,8)(7,16)(8,9)(9,9)(10,5)};
        \addplot+[ybar] plot coordinates {(2,5)(3,10)(4,12)(5,9)(6,12)(7,14)(8,13)(9,20)(10,12)};
      \end{axis}
      \begin{axis}[
          axis y line*=right,
          axis x line=none,
          ymin=0.5, ymax=1,
          ylabel=Entropy wrt. Pose Distribution,
        ]
        \addplot [mark=x, color=red, line width=0.5mm] coordinates
        {(2,0.941249672060794)(3,0.843261069972946)(4,0.808018384556546)(5,0.790800447319224)(6,0.781367523169684)(7,0.862244887939297)(8,0.805922238094384)(9,0.782618352676856)(10,0.771570589891531)};
      \end{axis}
    \end{tikzpicture}
  }
  \caption{$\MC$}
\end{subfigure}
\caption{
On \panoptic, for three AL strategies, we visualize the pose distribution of sampled frames. Colors represent different clusters, and the red curve tracks the entropy of poses (wrt. cluster IDs) over AL iterations.
$\MC$ produces diverse samples (higher entropy) and focuses more on under-represented clusters, leading to consistently superior performance.
}
\label{fig-pose-cluster}
\vspace{-0.2in}
\end{figure*}
\begin{figure}[t]
  \centering
  \resizebox{0.8\columnwidth}{!}{
      \input{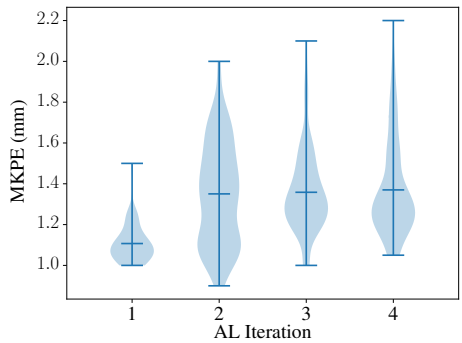}
  }
  \caption{
  Self-training (Alg.~\ref{alg-self-training}): the deviation between sampled
  pseudo-labels and corresponding ground truth, measured in MKPE, for \panoptic.
  Our selection strategy ensures that the pseudo-labels are accurate, and
  prevents ``drifting'' over time.
  }
  \label{fig-sal-dist}
  \vspace{-0.2in}
\end{figure}

\subsubsection{Data Augmentation, Self-Training, and AL}
The comparison between AL, AL$+\sal$, AL$+\textsc{RandAug}$, and
AL$+\sal$$+\textsc{RandAug}$ for different backbones and AL strategies are shown
in Fig.~\ref{fig-overall-comparison}. Data augmentation would improve the
efficiency of our AL based annotation process, especially at an earlier stage.
Larger performance gains can be observed on \random{} and PoseResNet-50 based AL
systems.

Self-training shows possible additional gains for all variations of experiments
with different AL strategies and data augmentation. However, performance
improvements from self-training saturates with higher performance models, \ie at
late stages of the AL process. Nonetheless, self-training can provide additional
gains ``for free'' in our AL based annotation process, since it does not incur
additional computational or annotation costs.

Finally, the choice of AL strategy outweigh data augmentation and self-training
in terms of the label-efficiency. Our proposed $\MC$ and $\CS$ would outperform
other compared ALs under all different setups.

\subsection{Ablation Studies}

\noindent\textbf{Diversity of Samples}:
We take one trial of our experiments with PoseResNet-50 on \panoptic{} where the
annotation amount is $50$ for each iteration, and study the distribution of
sampled poses. Intuitively, sampling more diverse poses (while still following
the data distribution) should help generalization. The ground truth 3D poses are
shifted to have keypoint 2 (waist) at origin, and clustered into 10 clusters
using K-means. We visualize the distribution of frames sampled by each AL
strategy based on this clustering in Fig.~\ref{fig-pose-cluster}, along with the
entropy computed from the discrete distributions.
The long-tail nature of the pose distribution can be seen from
Fig.~\ref{fig-pose-cluster}(a): samples from \random{} are unevenly distributed,
and dominated by clusters 1 and 8 in particular, which are common standing
poses. Compared to \random, the MPE strategy actually samples common pose
clusters more heavily, and frames from minority clusters (5, 7, 9) are
almost never sampled. In contrast, the proposed $\MC$, being based on an
uncertainty measure, attains much better pose diversity (higher entropy),
especially in the early iterations. This is because $\MC$ looks for
\emph{geometric} disagreements in the predictions, which are largely decoupled
from the prediction targets and their distribution.

\noindent\textbf{Accuracy of self-training pseudo-labels}:
The main challenge with pseudo-labels is to ensure their accuracy and avoid
drifting. In Fig.~\ref{fig-sal-dist}, we visualize the distribution of MKPE
between pseudo-labeled frames and their actual ground truth, over several AL
iterations. Our selection strategy maintains high accuracy ($<1.5$ mm MKPE on
average), resulting in consistent improvements over the course of AL.

\section{Conclusion}
In this paper, we propose an active learning framework for the data annotation
process of multi-view pose estimation. We first extend existing entropy-based
single-view AL strategies to multi-view, and then propose two AL strategies
utilizing 3D keypoint triangulation. The proposed CoreSet-Poses and Multi-View
Consistency strategies consistently outperform all AL and conventional
annotation baselines, for both body and hand pose estimation problems. In
addition, we introduce a self-training procedure using pseudo-labels, and
further improve the annotation efficiency with minimal cost. Our complete system
achieves state-of-the-art data annotation efficiency on \panoptic{} and \ih,
while using a fraction of the annotation cost and turn-around time.


{\small
  \bibliographystyle{ieee}
  \bibliography{bib}
}
\pagebreak


\appendix

\section{Supplementary Material}

This supplementary material provides details and additional ablation studies of
supporting experiments that are not presented in the main paper.  In the
following, we first present prior works on active learning for single-view human
pose estimation problem, which is discussed briefly in Sec. 3. Then, we present
the visualizations of the pose clusters used in Fig. 6 in the main paper.
Finally, we present our main results in the main paper (Fig. 3, 4 and 5) from a
different perspective to give a complete picture of the proposed methods.

\section{Prior Works on Single-view AL for Human Pose Estimation}
\begin{figure}
    \centering
    \includegraphics[width=.9\columnwidth, page=2, trim=0 30.7cm 54cm 0, clip]{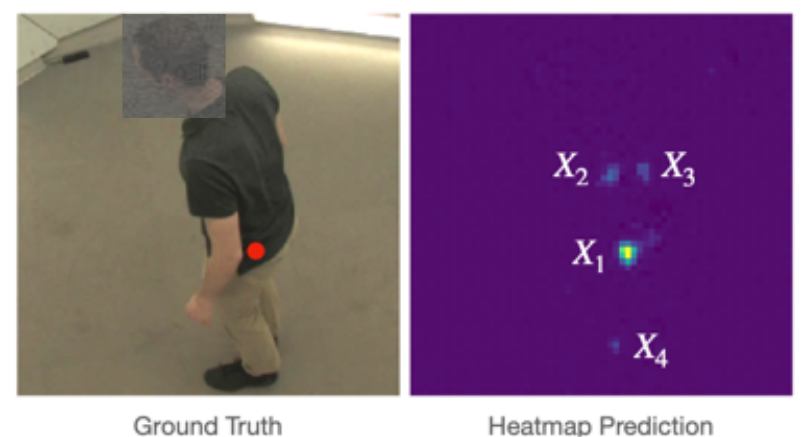}
    \caption{
      Illustration of entropy-based single-view AL strategies: Best vs. Second Best (BSB) and Multiple Peak Entropy (MPE).
      Let the normalized predicted heatmap for keypoint $p$ be $\hat{H}^p$, and $L^p = \{l^p_i\}$ be its local peaks.
      $\metric_\text{BSB} = \sum_{p}\hat{H}(l^p_2) - \hat{H}(l^p_1)$ and 
      $\metric_\text{MPE} = \sum_{p}\sum_{i=1}^4- \Pr(l^p_i) \log \Pr(l^p_i)$.
    }
    \label{fig-entropy}
  \end{figure}

The 2D heatmap representation in our setup for pose estimation naturally lends
itself to entropy-based formulations, since a heatmap encodes uncertainty in
the model's prediction, and can be normalized into a probability distribution
over the 2D grid using the softmax operator. For a predicted heatmap
$\heatmap^\kp$ of keypoint $k$, let $\localpeaks^\kp=\{\peak^\kp_1,\peak^\kp_2,
\ldots\}$ be a set of  2D coordinates of local peaks obtained by applying a
local maximum filter to $\hat{H}^k$, with $l_1^k$ being the argmax, and so on.
In the work of Liu and Ferrari~\cite{liu2017active}, several entropy-based
metrics are proposed, and a corresponding AL strategy is defined by sampling
top-scoring images under each metric. We now review these metrics. A visual
illustration of these metrics is shown in Fig.~\ref{fig-entropy}.

\subsection{Best vs. Second Best (BSB)}
The Best vs. Second Best metric~\cite{roth2006margin} is based on a margin
sampling idea, and defined as the difference between the top two local maximums
in the heatmap.  Intuitively, a smaller difference means larger uncertainty or
a multi-modal prediction.
\begin{equation}
  \bsb(V) = \frac{1}{\totalkp}
  \sum_{\kp=1}^\totalkp \left(\hat{\heatmap}^\kp(\peak^\kp_1) -\hat{\heatmap}^\kp(\peak^\kp_2) \right).
\end{equation}

\subsection{Multiple Peak Entropy (MPE)}
Multiple Peak Entropy is also introduced by Liu and Ferrari~\cite{liu2017active}
for single-view pose estimation.  The idea is that, as modes on a heatmap can be
spatially diffuse, simply comparing the highest and second highest would not be
able to differentiate between a single wide mode and multiple tight modes.
Instead, multiple peaks are considered together to better characterize the
uncertainty in a predicted heatmap.

To be concrete, MPE samples $\heatmap^p$ at all the local peaks $\localpeaks$,
and computes the resulting entropy:
\begin{equation}
  \mpe(V) = \frac{1}{\totalkp}
  \sum_{\kp=1}^\totalkp \sum_{\peak^\kp_i \in \localpeaks^\kp}-\Pr(\peak^\kp_i)\log\Pr(\peak^\kp_i),
\end{equation}
where
\begin{equation}
  \Pr(\peak^\kp_i) = \frac{\exp \heatmap^\kp(\peak^\kp_i)}
  {\sum_{\peak^\kp_j \in \localpeaks^\kp}\exp \heatmap^\kp (\peak^\kp_j)}.
\end{equation}
Note that the softmax operator is applied on the sparse set of local peaks
only.  Liu and Ferrari~\cite{liu2017active} found that MPE performs better over
the random baseline for single-view human pose estimation.

\begin{figure*}
  \centering
  \includegraphics[width=\textwidth, page=4, trim=0 20.7cm 9cm 0, clip]{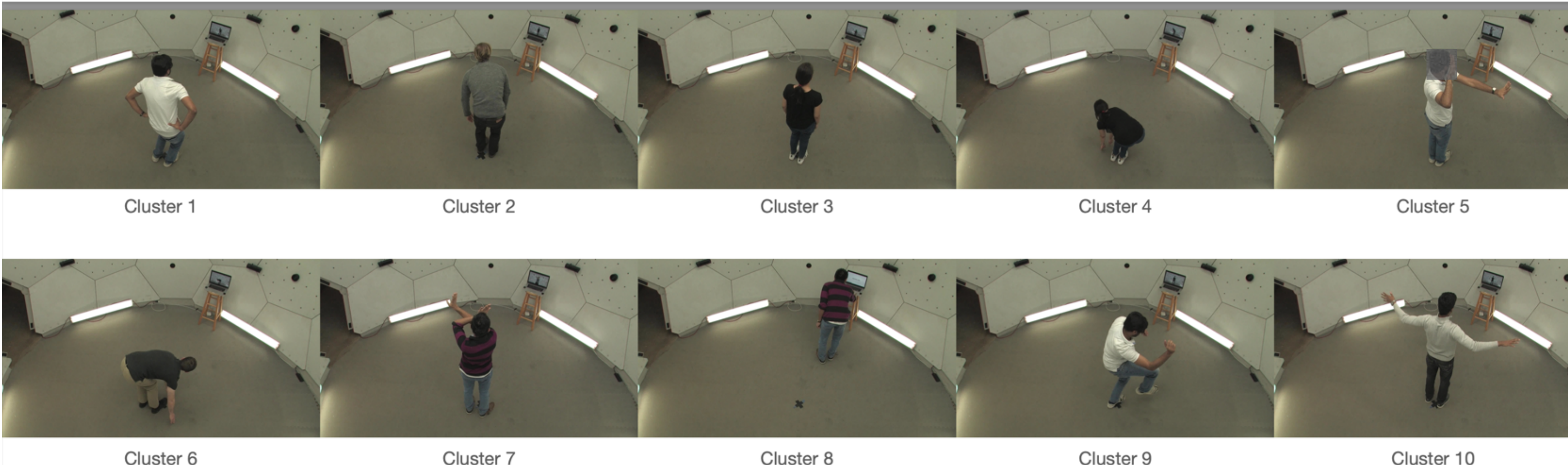}
  \captionof{figure}{
    Sample images from each of the 10 pose clusters (Fig 6 in the main paper),
    obtained by K-means.
  }
  \label{fig-visualization}
\end{figure*}

\begin{figure*}[h]
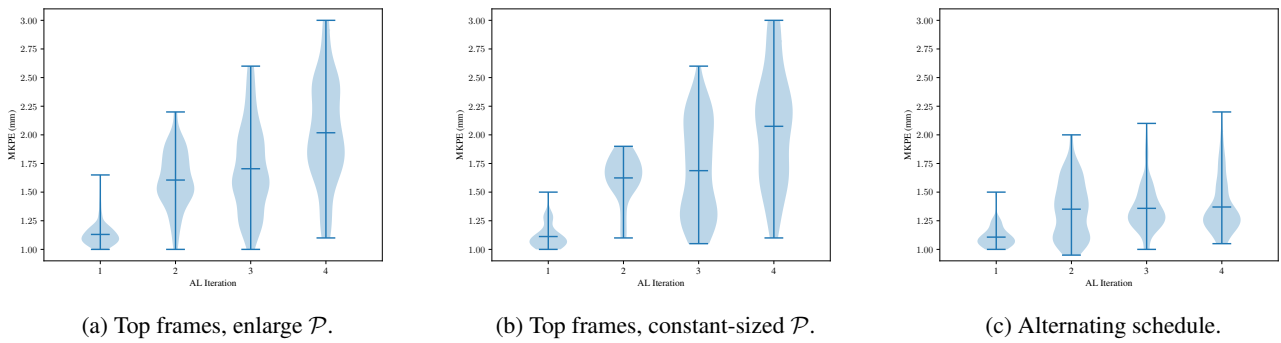

  \begin{subfigure}{0.33\textwidth}
    \centering
    \resizebox{\columnwidth}{!}{
      \input{figures/fig-sal-v1.pgf}
    }
    \caption{Top frames, enlarge $\pseudo$.}
    \label{fig-sal-v1}
  \end{subfigure}
  \begin{subfigure}{0.33\textwidth}
    \centering
    \resizebox{\columnwidth}{!}{
      \input{figures/fig-sal-v2.pgf}
    }
    \caption{Top frames, constant-sized $\pseudo$.}
    \label{fig-sal-v2}
  \end{subfigure}
  \begin{subfigure}{0.33\textwidth}
    \centering
    \resizebox{\columnwidth}{!}{
      \input{figures/fig-sal-v3.pgf}
    }
    \caption{Alternating schedule.}
    \label{fig-sal-v3}
  \end{subfigure}
  \caption{Comparisons between pseudo-labeling strategies: deviation between
    sampled pseudo-labels and corresponding ground truth, measured in MKPE.
    }
  \label{fig-sal-v1-v2-v3}
\end{figure*}

\begin{figure*}[h]
  \centering
  \begin{subfigure}{0.35\textwidth}
    \resizebox{\columnwidth}{!}{
      \begin{tikzpicture}
        \begin{axis}[
            grid=both,
            grid style={line width=.1pt, draw=gray!40},
            title = Panoptic / PoseResNet-50,
            ylabel= 3D MKPE (mm),
            xlabel= Percent of Dataset Labeled,
            legend pos = north east,
            xticklabels={0,0,$5\%$,$10\%$,$15\%$,$20\%$},
            legend style={font=\small, fill=white, fill opacity=0.6, draw opacity=1, text opacity=1},
            legend cell align={left},
            cycle list name=default,
            scaled x ticks = false,
            ymin=2.2,ymax=5.2
          ]
          \addplot table[x={PERCENT},y={AL-R-P-RESNET}, y error={AL-R-P-RESNET-ERROR}]{data/p.txt};
          \addplot table[x={PERCENT},y={AUG-R-P-RESNET}, y error={AUG-R-P-RESNET-ERROR}]{data/p.txt};
          \addplot table[x={PERCENT},y={AL-T-P-RESNET}, y error={AL-T-P-RESNET-ERROR}]{data/p.txt};
          \addplot table[x={PERCENT},y={AUG-T-P-RESNET}, y error={AUG-T-P-RESNET-ERROR}]{data/p.txt};
        \end{axis}
      \end{tikzpicture}
    } 
  \end{subfigure}
  \hspace{0.5in}
  \begin{subfigure}{0.36\textwidth}
    \resizebox{\columnwidth}{!}{
      \begin{tikzpicture}
        \begin{axis}[
            grid=both,
            grid style={line width=.1pt, draw=gray!40},
            title = Panoptic / HRNet,
            ylabel= 3D MKPE (mm),
            xlabel= Percent of Dataset Labeled,
            xticklabels={0,0,$5\%$,$10\%$,$15\%$,$20\%$},
            legend pos = north east,
            legend style={font=\small, fill=white, fill opacity=0.6, draw opacity=1, text opacity=1},
            legend cell align={left},
            cycle list name=default,
            scaled x ticks = false,
            ymin=2.2,ymax=4.0
          ]
          \addplot table[x={PERCENT},y={AL-R-P-HR}, y error={AL-R-P-HR-ERROR}]{data/p.txt}; 
          \addplot table[x={PERCENT},y={AUG-R-P-HR}, y error={AUG-R-P-HR-ERROR}, col sep=comma]{data/panoptic-1106.txt}; 
          \addplot table[x={PERCENT},y={AL-T-P-HR}, y error={AL-T-P-HR-ERROR}]{data/p.txt};
          \addplot table[x={PERCENT},y={AUG-T-P-HR}, y error={AUG-T-P-HR-ERROR}]{data/p.txt};
        \end{axis}
      \end{tikzpicture}
    } 
  \end{subfigure}
  \begin{subfigure}{\textwidth}
    \centering
    \resizebox{!}{1.4em}{
      \begin{tikzpicture}
        \begin{customlegend}[anchor=north west, legend columns=6,
            legend style={fill=none,draw=none,align=center,column sep=2ex},
            legend entries={\random,
              \random$+$\textsc{RandAug},
              \MC,
              \MC$+$\textsc{RandAug},
          }]
          \addlegendimage{mark=x, color=red}
          \addlegendimage{mark=x, color=blue}
          \addlegendimage{mark=x, color=cyan}
          \addlegendimage{mark=x, color=violet}
        \end{customlegend}
      \end{tikzpicture}
    }
  \end{subfigure}
  \caption{ 
    Effects of data augmentation using RandAugment on \panoptic. \MC{} achieves
    better label efficiency than $\random+\textsc{aug}$ without data
    augmentation. 
  }
  \label{fig-al-noaug-vs-aug}
\end{figure*}
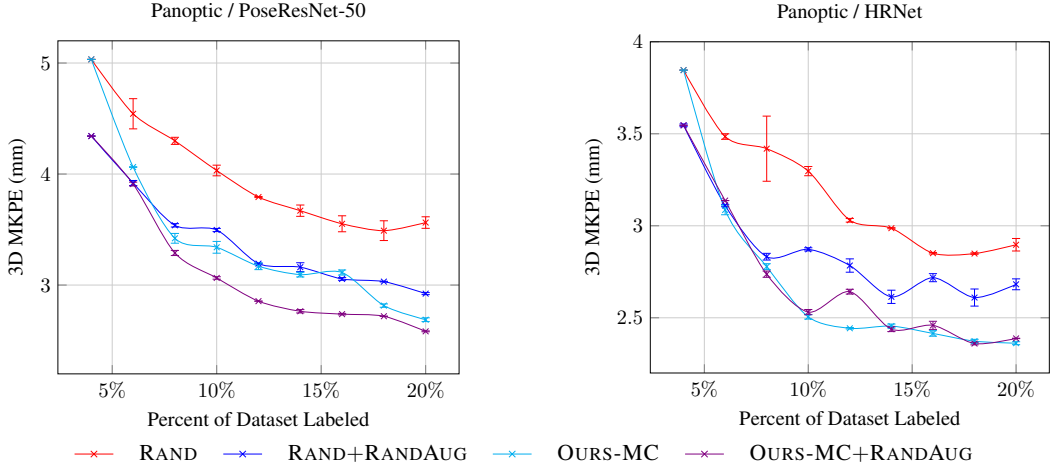
\begin{figure*}[h]
\centering
\begin{subfigure}{0.35\textwidth}
  \resizebox{\columnwidth}{!}{
    \begin{tikzpicture}
      \begin{axis}[
          grid=both,
          grid style={line width=.1pt, draw=gray!40},
          title = Panoptic / PoseResNet-50,
          ylabel= 3D MKPE (mm),
          xticklabels={0,0,$5\%$,$10\%$,$15\%$,$20\%$},
          legend pos = north east,
          legend style={font=\small, fill=white, fill opacity=0.5, draw opacity=1, text opacity=1},
          legend cell align={left},
          cycle list name=default,
          scaled x ticks = false,
          ymin=2.3,ymax=5.6,
        ]
        \addplot table[x={PERCENT},y={AL-R-P-RESNET}, y error={AL-R-P-RESNET-ERROR}]{data/p.txt};
        \addplot table[x={PERCENT},y={SALNOAUG-10-R-P-RESNET}, y error={SALNOAUG-10-R-P-RESNET-ERROR}]{data/p.txt};
        \addplot table[x={PERCENT},y={AL-T-P-RESNET}, y error={AL-T-P-RESNET-ERROR}]{data/p.txt};
        \addplot table[x={PERCENT},y={SALNOAUG-10-T-P-RESNET}, y error={SALNOAUG-10-T-P-RESNET-ERROR}]{data/sal-panoptics-triangulation.txt};
      \end{axis}
    \end{tikzpicture}
  }
\end{subfigure}
\hspace{0.5in}
\begin{subfigure}{0.36\textwidth}
  \resizebox{\columnwidth}{!}{
    \begin{tikzpicture}
      \begin{axis}[
          grid=both,
          grid style={line width=.1pt, draw=gray!40},
          title = Panoptic / HRNet,
          ylabel= 3D MKPE (mm),
          xticklabels={0,0,$5\%$,$10\%$,$15\%$,$20\%$},
          legend pos = north east,
          legend style={font=\small, fill=white, fill opacity=0.5, draw opacity=1, text opacity=1},
          legend cell align={left},
          cycle list name=default,
          scaled x ticks = false,
          ymin=2.2, ymax=4.3
        ]
        \addplot table[x={PERCENT},y={AL-R-P-HR}, y error={AL-R-P-HR-ERROR}]{data/p.txt};
        \addplot table[x={PERCENT},y={SALNOAUG-10-R-P-HR}, y error={SALNOAUG-10-R-P-HR-ERROR}]{data/p.txt};
        \addplot table[x={PERCENT},y={AL-T-P-HR}, y error={AL-T-P-HR-ERROR}]{data/p.txt};
        \addplot table[x={PERCENT},y={SALNOAUG-10-T-P-HR}, y error={SALNOAUG-10-T-P-HR-ERROR}]{data/p.txt};
      \end{axis}
    \end{tikzpicture}
  }
\end{subfigure}
\begin{subfigure}{\textwidth}
  \centering
  \resizebox{!}{1.4em}{
    \begin{tikzpicture}
      \begin{customlegend}[anchor=north west, legend columns=6,
          legend style={fill=none,draw=none,align=center,column sep=2ex},
          legend entries={$\random$,
            $\random+\sal$,
            $\MC$,
            $\MC+\sal$,
        }]
        \addlegendimage{mark=x, color=red}
        \addlegendimage{mark=x, color=blue}
        \addlegendimage{mark=x, color=cyan}
        \addlegendimage{mark=x, color=violet}
      \end{customlegend}
    \end{tikzpicture}
  }
\end{subfigure}
\caption{ 
  AL + self-training ($\sal$) on \panoptic{} \emph{without} data augmentation.
  X-axis: percent of dataset labeled. When combined with AL, our automated
  self-training strategy enables additional label efficiency gains at no extra
  computational cost, especially for \random{} and during the early stages of
  training. Best viewed in color. 
}    
\label{fig-sal-noaug}
\end{figure*}
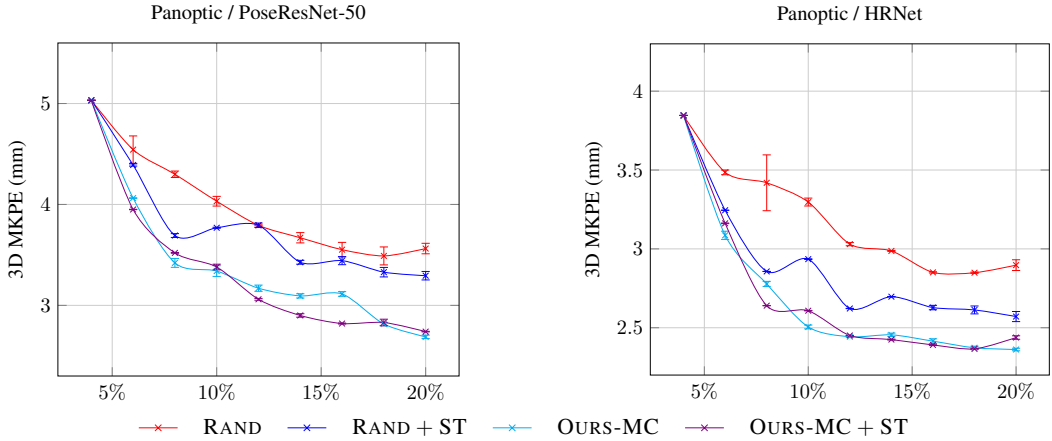

\subsection{Random}
Random sampling is a simple and very effective baseline strategy in active
learning for all kinds of tasks~\cite{mittal2019parting,viewal}. For pose
estimation, random selection of frames from $\unlabeled$ ensures that the
sampled poses closely follow the training distribution during the AL process.

\section{Visualization of 3D Pose}
As stated in our main paper, we study the distribution of sampled frames with
respect to a discrete clustering of ground truth poses.  In Fig.~6 of the main
paper, we have visualized the distribution of frames sampled by each AL
strategy, along with the entropy values computed from the discrete
distributions.  The ground truth 3D poses are shifted in 3D to have keypoint 2
(waist) at origin, and then we use K-means to cluster them into 10 clusters.
Sample images from each cluster are visualized in Fig~\ref{fig-visualization}.
The visualization confirms the findings that the proposed \MC{} samples frames
with better diversity in poses (higher entropy), especially in the early
iterations. 

\section{Ablation Studies on Self-Training and Augmentation}

\subsection{Ablation Studies on Self-Training}
In addition to the differences between the proposed self-training algorithm and
the multi-view bootstrapping method~\cite{simon2017hand} mentioned in the main
paper, self-training produces new and more accurate pseudo-labels as the amount
of human-annotated data increases with the AL iteration.  Here, we detail the
design choices for our specific self-training strategy.

We have considered the following three strategies.  In
Fig.~\ref{fig-sal-v1-v2-v3}, we plot the distributions of MKPE between
pseudo-labels in $\pseudo$ and their corresponding ground truth, over the first
four AL iterations.
\begin{enumerate}
  \item 
    Fig.~\ref{fig-sal-v1}. Enlarge $\pseudo$ in each AL iteration with the top
    pseudo-labeled frames. (Selection criterion is discussed in the main paper.)
  \item 
    Fig.~\ref{fig-sal-v2}. Keep the size of $\pseudo$ constant, and pick the
    top pseudo-labeled frames in each AL iteration.
  \item 
    Fig.~\ref{fig-sal-v3}. Alternating schedule (described in the paper): in
    each AL iteration $i$, pick top frames that are not already in
    $\pseudo_{i-1}$ from the last iteration, to form $\pseudo_i$ for the
    current iteration.
\end{enumerate}

To begin with, the first strategy is easily susceptible to label drifting, as
more and less accurate pseudo-labels would enter $\pseudo$ and pollute the
training set over time.  Somewhat surprisingly, the second strategy of keeping
the size of $\pseudo$ constant does not work either.  We have empirically
verified that, in this scenario, the set of frames selected to form  $\pseudo$
is very stable across iterations.  Then, with every passing iteration, this
strategy essentially re-labels a same set of frames, using a new model trained
on a training set containing them, and the errors would accumulate.  Note that
in this case, the model needs to achieve zero training error on $\pseudo_i$ in
every AL iteration $i$  for the pseudo-labels to remain the same, let alone
improve.

Lastly, we found the alternating schedule to be robust against label drifting.
In each iteration $i$, all frames in $\pseudo_{i-1}$ are evicted, and prevented
from re-entering until the next iteration.  This effectively avoids the above
error accumulation problem, as a model trained on frames from $\pseudo_i$
(among others) is never used to infer pseudo-labels on the same set of frames.

\subsection{Ablation Studies on Data Augmentation}
We present the effect of RandAugment on \panoptic{} in
Fig.~\ref{fig-al-noaug-vs-aug} that is presented separately in Fig.~4 and
Fig.~5 in the main paper.  For each training image, we randomly apply two of
the following augmentation operations: 

\begin{itemize}
  \item \texttt{Rotate} (within $\pm 30^\circ$)
  \item \texttt{AutoContrast}
  \item \texttt{Equalize}
  \item \texttt{Invert}
  \item \texttt{Posterize}
  \item \texttt{Solarize}
  \item \texttt{Color}
  \item \texttt{Contrast}
  \item \texttt{Brightness}
  \item \texttt{Sharpness}
\end{itemize}

In the case of image rotation, we also rotate the target heatmap by the same
amount. All other operations are label-preserving and do not alter the heatmap.

Overall, we find that the choice of AL strategy outweighs data augmentation.
For example, \MC{} without data augmentation even outperforms
$\random+\textsc{RandAug}$ with PoseResNet-50 backbone. For \MC{} with the
HRNet backbone, the performance gain from data augmentation gets smaller as it
saturates more quickly towards the fully-supervised baseline.

Additionally, we compare performances of self-training on \random{} and \MC{}
without RandAugment and present the comparison in Fig.~\ref{fig-sal-noaug} to
complement Fig. 5 in the main paper. Self-training suffers from the fact that
no augmentation is used in these experiments and only provides marginal gains,
with one exception to \random{} with HRNet,  where self-training shows a
slightly larger gain.

\end{document}